\documentclass{nle}

\DeclareMathAlphabet{\mathpzc}{OT1}{pzc}{m}{it}
\usepackage{amsmath}
\usepackage{graphicx}
\usepackage{tabularx}
\usepackage{natbib}
\usepackage{todonotes}
\ifpdf%
\usepackage{epstopdf}%
\else%
\fi
\usepackage{multirow}
\usepackage{pgfplots}
\usetikzlibrary{patterns}
\usepackage{siunitx}
\usepackage{float}

\definecolor{lightcornflowerblue}{rgb}{0.6, 0.81, 0.93}
\definecolor{cadmiumorange}{rgb}{0.93, 0.53, 0.18}
\definecolor{asparagus}{rgb}{0.53, 0.66, 0.42}
\definecolor{brandeisblue}{rgb}{0.0, 0.44, 1.0}
\definecolor{deepcarrotorange}{rgb}{0.91, 0.41, 0.17}
\definecolor{lavender}{rgb}{0.71, 0.49, 0.86}
\definecolor{darklavender}{rgb}{0.45, 0.31, 0.59}
\definecolor{sapgreen}{rgb}{0.31, 0.49, 0.16}
\definecolor{bananayellow}{rgb}{1.0, 0.88, 0.21}
\usepackage{url}

\pgfplotsset{
    compat=newest,
    scaled y ticks=false,
    yticklabel style={
        /pgf/number format/fixed,
        /pgf/number format/precision=5
    }
}

\newcommand{\KnowMAN}{\textsc{KnowMAN}}
\newcommand{\bKM}{\textsc{blind KnowMAN}}
\newcommand{\MV}{\textsc{Majority Vote}}
\newcommand{\Snorkel}{\textsc{Snorkel-DP}}
\newcommand{\XPASC}{\textsc{XPASC}}
\newcommand{\chis}{\textsc{Chi Square}}
\newcommand{\PPMI}{\textsc{PPMI}}
\newcommand{\wea}{\textsc{WeaSEL}}

\usetikzlibrary{arrows}
\usetikzlibrary{shapes}

\begin{document}
\label{firstpage}

\lefttitle{\XPASC{}, M\"{a}rz et.al}
\righttitle{\XPASC{}, M\"{a}rz et.al}

\papertitle{Article}

\jnlPage{1}{00}
\jnlDoiYr{2022}
\doival{10.1017/xxxxx}

\title{\XPASC{}: Measuring Generalization in Weak Supervision by Explainability and Association}

\begin{authgrp}
\author{Luisa M\"{a}rz$^{\diamond,\ast,\star}$}
\author{Ehsaneddin Asgari$^{\ast}$}
\author{Fabienne Braune$^{\ast}$}
\author{Franziska Zimmermann$^{\circ}$}
\author{ Benjamin Roth$^{\diamond, \dag}$}

\affiliation{$^\diamond$ Research Group Data Mining and Machine Learning,\\  
                        Faculty of Computer Science, University of Vienna,\\
                        Vienna, Austria\\
        }

\affiliation{$^\dag$ Faculty of Philological and Cultural Studies, \\
                        University of Vienna,\\
                        Vienna, Austria\\
}

\affiliation{$^\ast$ AI Innovation \& Pre-Development, \\
                    Data:Lab, Volkswagen AG,\\
                    Munich, Germany\\
\email{luisa.k.maerz@gmail.com}\\
}

\affiliation{$^\star$ UniVie Docotoral School Computer Science,\\
            Vienna, Austria\\}

\affiliation{$^\circ$ CAPE Analytics, \\
Mountain View, California, United States
}

\end{authgrp}

\history{(Received 06 May 2022; revised 22 November 2022; accepted xx xxx xxx)}
%\received{20 March 1995; revised 30 September 1998}

\begin{abstract}
Weak supervision is leveraged in a wide range of domains and tasks due to its ability to create massive amounts of labeled data, requiring only little manual effort. 
Standard approaches use labeling functions to specify signals that are relevant for the labeling.
It has been conjectured that weakly supervised models over-rely on those signals and as a result suffer from overfitting.
To verify this assumption, we introduce a novel method, \XPASC{} (eXPlainability-Association SCore), for measuring the generalization of a model trained with a weakly supervised dataset.
Considering the occurrences of features, classes and labeling functions in a dataset, \XPASC{} takes into account the relevance of each feature for the predictions (explainability) of the model as well as the connection of the feature with the class and the labeling function (association), respectively.
The explainability is measured using occlusion in this work.
The association in \XPASC{} can be measured in two variants: \XPASC{}-\chis{} measures associations relative to their statistical significance, while \XPASC{}-\PPMI{} measures association strength more generally.
 
We use \XPASC{} to analyze \KnowMAN{}, an adversarial architecture intended to control the degree of generalization from the labeling functions and thus to mitigate the problem of overfitting. 
On one hand, we show that \KnowMAN{} is able to control the degree of generalization through a hyperparameter. 
On the other hand, results and qualitative analysis show that generalization and performance do not relate one-to-one, and that the highest degree of generalization does not necessarily imply the best performance. 
Therefore methods that allow for controlling the amount of generalization can achieve the right degree of benign overfitting.
Our contributions in this study are 
i) the \XPASC{} score to measure generalization in weakly-supervised models, 
ii) evaluation of \XPASC{} across datasets and models and 
iii) the release of the \XPASC{} implementation.

\end{abstract}

\maketitle

\section{Introduction}
Many machine learning architectures still require large amounts of labeled training data, resulting in static data sets with limited usability for changing data distributions or task definitions. 
%Such large annotated corpora are a scarce resource, especially in real-world scenarios. 
Manual annotation is both expensive and time-consuming and thus not always practically feasible or convenient.
One way to circumvent this problem is to use weak supervision. 
Weak supervision methods utilize different knowledge sources such as knowledge bases, heuristics, or taxonomies to annotate large amounts of data automatically. 
The knowledge of the external sources is encoded in  \emph{labeling functions}, programmatically specified heuristics (e.g., keywords, patterns, database lookups) that trigger the automatic annotation of a specific output.
A labeling function can also be thought of as a decision rule that is defined based on prior (expert) knowledge.
If this rule is matched, the appropriate output is assigned to the instance.
Due to the fact that labeling functions only consider a narrow context for triggering an annotation, it is likely that some weak labels are noisy and/or imprecise.
Moreover, large sets of weakly annotated instances all follow similar patterns (captured by the same labeling function), and it has been conjectured that weakly supervised models rely too heavily on the labeling functions and therefore suffer from overfitting \citep{Mrz2021KnowMANWS,Dehghani2017AvoidingYT}.

Approaches to tackle the problem of noisy weak labels can be categorized into two main groups: 
Those who try to filter out the noisy labels for training \citep{Ren2020DenoisingMW,Sukhbaatar2014TrainingCN,Dehghani2017AvoidingYT} and those who try to estimate the accuracy of the labeling functions or the weak sources \citep{Fu2020FastAT,snorkel}.
See \citet{Zhang2022ASO} for a detailed survey on weak supervision approaches and labeling function modeling.
None of them addresses the problem of overfitting to (or, inversely, generalization from) labeling functions.
As an alternative, we developed \KnowMAN{} \citep{Mrz2021KnowMANWS}, an adversarial architecture with the objective to shift the focus for the learned representation of a model away from the labeling functions towards a more general representation. 
This is achieved through a hyperparameter that controls the influence of the labeling functions on the feature representation.
\KnowMAN{} has explicitly been designed to overcome the problem of overfitting to the noisy labeling function signals. 
However, while training in the \KnowMAN{}-settings increases the prediction quality for the studied weakly supervised neural networks, it could not be \emph{directly} evaluated whether the \KnowMAN{} actually increases generalization from the labeling functions.

In this study, we present \textbf{\XPASC{} (eXPlainability-Association SCore)} to observe generalization from noisy signals in weakly supervised models more closely.
The intuition behind the score is that models suffering from overfitting to labeling functions have a low ability to generalize, and will heavily rely on features associated with labeling functions for prediction.
Generalization in the scope of this work means the capability of a model to abstract from the labeling function signals and to learn representations based on various signals and parts of the input.
A higher generalization should ultimately lead to the representation being more robust against misleading labeling functions and being able to represent the input with as many aspects as possible. Accordingly, a greater generalization from the weak source indicates a smaller degree of overfitting.
In this work we consider the strongly information-carrying surface forms, i.e. the individual tokens, of an instance as features.
By using \XPASC{} the generalization ability of a model, given a data set, can be measured. 
The score combines the relevance of each feature for the prediction of a weakly supervised model (explainability) with the connection of the feature with the class and the labeling function (association).

To measure prediction relevance of the single features we leverage a method from XAI (eXplainable AI), namely occlusion. In general, XAI methods aim to give insights into the output of machine learning models to make internal processes more transparent to users. Taking into account the explainability method is central to the analysis with XPASC, and also represents an unconventional use case of XAI.

To compute association in \XPASC{} we propose two different methods: \XPASC{}-\chis{} which relies on statistical association strength, and \XPASC{}-\PPMI{} which measures the more general information-theoretic association strength.

Apart from introducing the formal details of \XPASC{}, we study \KnowMAN{} and \wea{} as well as two traditional weak supervision approaches, \MV{} and \Snorkel{} (Data Programming), with respect to their generalization, as measured by \XPASC{}.
Our findings show that the \KnowMAN{} architecture is able to control the degree of generalization in direct relation to its hyperparameter $\lambda$.
In fact, \KnowMAN{} can get the model to focus more on words that are associated with the class (generalize more), and even ignore features highly associated with single, misleading labeling functions.

We have observed that the generalization of a model and its performance are not one-to-one related, and that at a certain point there can be too much generalization from the weak signals.
Our observations show that the two different association computations, \PPMI{} and \chis{}, behave analogously in the overall picture.
However, we find that the values of the association based on \PPMI{} are distributed across a larger space. 
This means that \PPMI{} also takes into account the "long tail" of data sets and presents the reality in the data more straightforwardly.
The values of the \chis{}-based association, on the other hand, have a denser distribution. 
So \chis{} also appears to be more resilient to outliers in the data.

Our main contributions in this paper are:
\begin{itemize}
    \item the proposal and detailed introduction of \XPASC{} to measure generalization from weak signals
    \item the evaluation of \XPASC{} across models and data sets
    \item the confirmation of the hypothesis and functionality of \KnowMAN{}
    \item the release of the \XPASC{} implementation\footnote{\url{https://github.com/LuisaMaerz/XPASC}}. 
\end{itemize}

The remainder of the paper is structured as follows: 
After investigating related work in the fields of weak supervision and overfitting metrics we describe the method the \XPASC{} formally. Section \S\ref{sec:exp} gives an overview over models and data sets used in this work. The analysis is divided in quantitative and linguistic results and followed by the conclusion. 
\section{Related Work}
We consider the concepts of overfitting and generalization to be related in that overfitting can be a result of low generalization. 
In the weak supervision context, this means that a model that abstracts little from the labeling functions, i.e. has low generalization, is more likely to overfit to these misleading signals.
In the following we present approaches that focus either on overfitting or on generalization.
Some are tailored to weak supervision, while others deal with overfitting and generalization in natural language processing in general.

Overfitting of machine learning models has been repeatedly identified as a problem. 
In machine learning generally, overfitting means that a model has adapted too much to the training data and can therefore no longer perform well on newly seen data.
Model performance on unseen data or held-out test sets is therefore typically used as an indicator for overfitting. 
If the result on unseen data is much worse than on training data it is likely that the model overfitted to the training data.
\cite{salman2019overfitting} analyze the training dynamics and the application of models to unseen data to observe overfitting.
Other works measure overfitting by the total number of parameters, where a low number of parameters indicates less overfitting than a high number of parameters. %CITATION??
\citep{dloverfitting} measure overfitting through the number of parameters of a model and its accuracy on the test set. 

%When it comes to weak supervision, overfitting to noisy input data has been addressed in the past as well. 
\cite{measuringoverfitting} analyze overfitting caused by test set reuse on a large set of Kaggle competitions. 
The assumption is that if many models are centered towards one test set, overfitting of the models to that test set is likely.
However, with their experiments they show that there is no significant overfitting due to test set reuse in Kaggle. 
\cite{measuringoverfitting} provide a simple measurement for the adaptivity gap between the losses on train and test set. 
Here they use the fact that Kaggle provides two types of test sets with which this gap can be approximated very well.
Unlike us, they cover overfitting to test set reuse rather than overfitting to labeling functions.

Due to the nature of weak supervision, models may overfit to systematic errors and biases introduced by the automatic labeling process.
\cite{yu2020fine} fine tune a pre-trained language model with weak supervision.
This is challenging, because large language models have a higher risk to overfit due to their large amount of parameters anyways. 
Errors are more propagated due to overfitting, which degrades performance and prevents the models from learning properly.
\cite{yu2020fine} tackle this issue by contrastive self-training, what can be considered as denoising in the first place and reduces error propagation and overfitting to the noise. 
In their experiments, they use RoBERTa \citep{liu2019roberta} and fine-tune a simple classification head.
Like us, they use \MV{} (\textit{exact match} in their work) and \Snorkel{} for weak labeling.
However, they do not specifically address the impact of overfitting to labeling functions.
As in our previous work with \KnowMAN{} \citep{Mrz2021KnowMANWS}, their model aims at learning better representations from weakly labeled data. 
Unlike us, they use a contrastive approach that pulls labels with similar weak supervision signals closer together and pushes others further away in the feature space, rather than an adversarial network as in \KnowMAN{}. 
In recent work \citet{Zhang2022UnderstandingPW} propose the source-aware Influence Function to understand programmatic weak supervision. 
By observing changes in the loss of a model while utilizing the source-aware influence function, they gain insights in the influence of single data points, labeling functions or weak sources on model performance.
Like us, they aim to identify important parts of the input to make the model output more explainable. 
Similar to \KnowMAN{}, they try to reduce the impact of misleading labeling functions on model training. 
Although they provide some insight into what influences the training through the source-aware influence function, they do not use this information to compare different models, which is different from our work.
 
Generalization refers to a model's ability to perform well on unseen data, i.e., a model generalizes well if it overfits only slightly. 
For example \cite{ratner2019training} consider generalization of weak supervision sources observable through the estimation error of their trustworthiness. 
They claim that the generalization error scales with the number of unlabeled data points and try to minimize the loss for predicting weak labels without loss of generality.
By connecting generalization to the estimation error, generalization is not only observable, but also controllable. 
Like our metric, this can be viewed as a formal measurement of generalization.
Unlike our metric, their measure is tightly coupled with their specific weak supervision approach and not generally applicable as a universal tool to compare generalization across models and data sets. 

Many weak supervision approaches try to overcome a lack of generalization by denoising the weak sources \citep{Ren2020DenoisingMW, hsieh2022nemo}.
In contrast to these approaches, we address the issue of generalizing from weak supervision sources, instead of denoising them.
\cite{Fu2020FastAT} provide a weak supervision framework to model and label data by leveraging different weak supervision sources. 
In addition, they provide a bound for generalization. 
To do so, they measure the performance gap between the end model parametrization using outputs of the label model and the optimal end model parametrization over the true distribution of labels. 
%Unlike us, they do not take into account misleading labeling functions to measure generalization.
More efforts can be mentioned in studying generalization in general, e.g., measuring number of required ``strong'' labels~\citep{robinson2020strength},  studying generalization in algorithmic datasets~\citep{power2022grokking}, generalization in generative models by measuring the uniqueness of generated sample~\citep{mauri2022evaluating}.

Although there is work on both overfitting and generalization, and both are considered to be important issues in weak supervision, to the best of our knowledge \XPASC{} is the first universal score measuring overfitting to and generalization from labeling functions in weakly supervised models. Moreover, since our approach is based on explainability methods, it makes transparent which features are mainly responsible for overfitting or generalization.

\begin{figure}[t!]
    \centering
    \includegraphics[width=\textwidth]{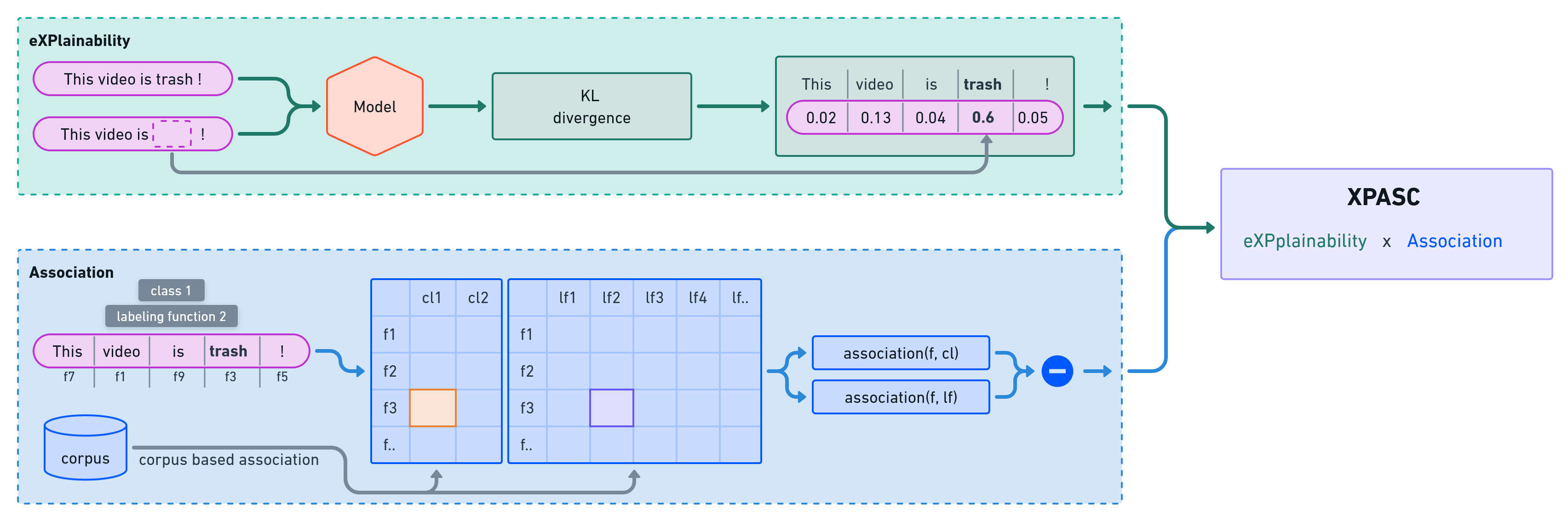}
    \vspace*{2.2cm}
    \caption{Modules of \XPASC{}:  product of \textit{eXPlainability} and \textit{Association}. 
    The \textit{eXPlainability} of a feature of an instance is calculated by the KL divergence of the class predictions for the entire instance and the instance without the respective feature. For each instance the explainability for all features is computed and the most important feature can be obtained.
    The \textit{Association} for an instance given a feature is calculated as the difference of the association of the feature with the class and with the labeling function, which in practice is a matrix lookup. The matrices are computed in advance and are based on coocurrences of features, classes and labeling functions in the data corpus.}
    \label{fig:XPASC}
\end{figure}

\section{The Explainability-Association Score} % for Measuring Generalization}
The goal of \XPASC{} is to measure generalization from weak signals for weakly supervised models.
The intuition behind \XPASC{} is that input parts or features that are most important for the class prediction of these models are highly associated with the heuristics used for annotating the training data. 
This is due to the fact that the model relies too much on the labeling functions and therefore tends to ignore other valuable signals for classification.

\XPASC{} measures to what degree a model, trained with a weakly labeled data set, can generalize from the information associated with the features, classes and labeling functions present in the data.
Several considerations such as how important features\footnote{In this work, by features we mean the observable tokens of an input instance, in contrast to learned features from a network.} are for the classification and how much features, classes and labeling functions are correlated are taken into account. 
Therefore \XPASC{} is composed of three parts:
i) the explainability of each feature for a model, 
ii) its association with the class, and
iii) its association with the labeling function. 
Note that both explainability and association contribute equally to \XPASC{}.
To calculate explainability, we use occlusion \citep{zeiler2014visualizing}. 
The association strength is measured either with the \PPMI{} or the \chis{}-score.
See Figure \ref{fig:XPASC} for an illustration of \XPASC{}.

\subsection{Explainability}
The term explainability expresses how important a single feature is for the classification of an instance, i.e., how the class prediction changes if the feature is omitted, masked or changed.
To determine the importance of each feature of an instance for the classification task, we use the explainability method of occlusion. The idea for occlusion originally came from computer vision proposed by \cite{zeiler2014visualizing}, and since then it has been also used for Natural Language Processing, e.g., by \cite{harbecke2020considering}, or \cite{ancona2017towards}. 

We perform occlusion in four steps as follows: we pass \textbf{i)} our input instance through our model and \textbf{ii)} the instance without the feature through the model. Explainability is then computed by \textbf{iii)} retrieving the prediction probabilities from both, and \textbf{iv)} calculating the Kullback-Leibler-Divergence \citep{kullback}:

\begin{equation}
    D_{KL}(P||Q) = \sum_{x \, \in \, X } P(x) \, log \left ( \frac{P(x)}{Q(x)} \right )
\end{equation}

where $P$ and $Q$ are discrete probability distributions and $X$ is a shared probability space.
By computing how different two probability distributions are, the Kullback-Leibler-Divergence indicates how much the occlusion affects the prediction in our case.
Accordingly, we define the explainability of an instance and a feature as:

\begin{equation}
    S_{xp}(i,f) = D_{KL} \left ( P(i)||Q(i \setminus f) \right ) 
\end{equation}

where $P(i)$ is the model prediction for the entire instance and $Q(i \setminus f)$ is the model prediction for the instance with feature $f$ omitted. 
Note that both predictions are vectors of probability distributions over the set of possible classes.  
The smaller this difference of the two probability distributions, the less important the feature is for the classification result. 
In any case, the Kullback-Leibler-Divergence is between zero and one.
The highest explainability is assigned to the features where the two prediction distributions differ the most, i.e., where the prediction changes greatly if the feature is omitted.

\subsection{Association}
Association measures the degree of correlation between observations.
To find out how much a feature is correlated with its class and with its labeling function we calculate two association matrices with the following shapes:
\[
| \textbf{C} | = classes \times features
\]
\[
| \textbf{L} | = labeling\_functions \times features
\]
\indent where $\textbf{C}$ is the association matrix for classes and $\textbf{L}$ is the matrix for labeling functions. 
The details for the matrix calculation are explained in section \ref{sec:chis} and section \ref{sec:ppmi}.
During the calculation of \XPASC{}, the respective association value is looked up in the matrices for each feature given its class and its labeling function. 
To put both associations (feature and class/ feature and labeling function) in relation we subtract their scores and arrive with the overall association for a feature given its instance:

\begin{equation}
S_{asc}(i,f) = \sum_{j=1}^{N} \left ( \textbf{C}_{c^i f} - \textbf{L}_{l^i_j f} \right )
\label{formula:asc}
\end{equation}

where we iterate over all matching labeling functions $l$ for instance $i$. 
Variable $f$ denotes the feature, $\textbf{C}_{c^if}$ is the value of the association-matrix with respect to the class label of instance $i$, $c_i$, and $\textbf{L}_{l^i_j f}$ the value for the $j$'th matching labeling function, $l^{i}_y$. 
By computing association that way the result can either be positive, negative or zero. 
Depending on the exact result, the score can then be interpreted directly.
The larger (positive) $S_{asc}$, the more feature $f$ is associated with the class, 
the smaller (negative) $S_{asc}$, the more feature $f$ is associated with the labeling function. 
The closer the score is to zero, the more similar the association of the feature with the class and the labeling function. 

Association can be modeled in different ways. 
We calculate it in two manners: using i) a chi squared test or ii) the positive pointwise mutual information.

\subsubsection{\chis{}-based association} \label{sec:chis}
For this option we calculate the association matrices based on univariate feature selection. 
This works by selecting the best features based on univariate statistical tests, in our case a chi squared ($\chi^2$) test. 

\begin{equation}
    \chis{}(z,f) = \, \frac{\left ( o_{zf} - \tilde{o}_{zf} \right )^2}{\tilde{o}_{zf}}
    \label{generalchi}
\end{equation}

where $f$ is the feature, $z$ represents a label (either class label or labeling function), $o_{zf}$ is the absolute frequency of the combination of class/ labeling function and a feature (observed value) and $\tilde{o}_{zf}$ is the expected value of the absolute frequency of the combination of class / labeling function and a feature.

The $\chi^2$ test measures the dependence between the features and the class/labeling function. 
Features with a high \chis{} score are likely to be independent of the class/labeling function and therefore more irrelevant for the classification. 
The smaller the \chis{} result, the more a feature is related to the class/labeling function and, consequently, the more important it is.
Note that the \chis{} association expresses which of the features are most associated with the class and include positive as well as negative correlation. 
Thus also negative examples can be found among the most associated ones, e.g. ``{}brother''{} can have a very high association with the ``{}married to''{} relation, although it is a negative indicator for that relation. 

Formulas \ref{Cchi} and \ref{Lchi} define how to calculate one matrix entry for a feature and its corresponding class/ labeling function.

\begin{equation}
    \textbf{C}_{cf}^{\,\chis{}} \, = \, \chis{}(c,f)
    \label{Cchi}
\end{equation}

\begin{equation}
    \textbf{L}_{lf}^{\,\chis{}} \, = \, \chis{}(l,f)
    \label{Lchi}
\end{equation}

where $c$ is the class, $l$ the labeling function and $f$ the feature.

\subsubsection{\PPMI{}-based association} \label{sec:ppmi}
As a second option we calculate the positive pointwise mutual information (Equation \ref{PPMI}), where only the positive results of the pointwise mutual information (Equation \ref{PMI}) are taken into account. 
Assuming independence of two variables (in our case: a feature and a class/labeling function) PMI quantifies the discrepancy between the probability of their coincidence given their individual distributions and their joint distribution.

\begin{equation}
PMI(f,z) = log  \left ( \frac{P(f,z)}{P(f)P(z)} \right )
\label{PMI}
\end{equation}

\begin{equation}
    \PPMI{}(f,z) = \left\{\begin{matrix}
 PMI, \, & \text{if} \, PMI > 0  \\
 0 & else  \\
\end{matrix}\right.
\label{PPMI}
\end{equation}

where $f$ is a feature, $z$ a label (either class label or labeling function), $P(f,z)$ the joint probability of a feature and a label, $P(f)$ the probability of the feature and $P(z)$ the probability of a label.
Formulas \ref{CPPMI} and \ref{LPPMI} define how to calculate one matrix entry for a feature and its corresponding class/ labeling function.

\begin{equation}
    \textbf{C}_{cf}^\PPMI{} =  \PPMI{}(f,c)
\label{CPPMI}
\end{equation}

\begin{equation}
    \textbf{L}_{lf}^\PPMI{} =  \PPMI{}(f, l)
\label{LPPMI}
\end{equation}

where $c$ is the class, $l$ the labeling function and $f$ the feature.

\subsection{The Combined Score: \XPASC{}}
\XPASC{} (eXPlainability Association SCore) combines both explainability and association for one data set and a model. 
It measures how important each feature of an instance is and if it is more correlated with the class or with the labeling function.
By multiplying the two measures and summing up over all instances and features we obtain:

\begin{equation}
    S_{\XPASC{}}(d, m) = 1 + \left (
\frac{1}{N \times M} \sum_{i=1}^N \sum_{f=1}^M  S_{xp}(i,f) \times S_{asc}(i,f) \right )
\end{equation}

where $d$ is the data set, $m$ is the pre-trained model, $N$ is the number of instances and $M$ is the number of features.
To make sure that \XPASC{} is comparable across models and data sets we normalize by the data set size (number of instances times the number of features).
Multiplying both components gives small negative values, so we add one to the result to make the final \XPASC{} value above zero.
The multiplication allows to put the importance of a feature in relation to its association with the class and the labeling function.

The sharpness of the explainability measure could be increased by a temperature hyperparameter $\gamma$ for putting the focus only on the most relevant features ($\gamma\rightarrow\infty$) or equally on all features ($\gamma\rightarrow0$), changing the \XPASC{} formula as follows: 
$\left ( S_{xp}(i,f)^\gamma  \right ) \times \left ( S_{asc}(i,f) \right )$.
In this work, we considered the unchanged importance as given by the explainability algorithm ($\gamma=1$).

%BR: the following does not make sense - please leave out.
%Note that both scores, explainability and association, contribute equally to the final formula and are thus equally important.
%Should one wish to give more weight to either explainability or association, this could be solved via a hyperparameter that would change the multiplication in the \XPASC{} formula as follows: 
%$\left (  \gamma \times S_{xp}(i,f) \right ) \times \left ( (1-\gamma) \times S_{asc}(i,f) \right )$.
%In this work we considered both components to be of equal importance.

Thus, 
%the score measures how many the important elements (those that are effectively used for prediction) are associated with the class label rather than the labeling function, i.e., 
a high \XPASC{} indicates that many of the most important features (those that are effectively used for prediction) are correlated with the class instead of the labeling function. 
We can conclude that a high \XPASC{} indicates more independence from the labeling functions and accordingly a greater generalization from the weak source. 
This also indicates a smaller degree of overfitting to the weak signals.
%Note that the numbers of the \XPASC{} results are very small because the score is normalized by the totality of features and instances. 
Note that the results of \chis{}-based \XPASC{} and \PPMI{}-based \XPASC{} are scaled differently. 
This is due to their different characteristics, as well as the specific result values of the two calculations.

\section{Weak Supervision Methods and Datasets} \label{sec:exp}
For our experiments we study four different weak supervision approaches, \KnowMAN{}, \MV{}, \Snorkel{} (Data Programming) \citep{snorkel} and \wea{} \citep{weasel}. 
Of those, only \KnowMAN{} provides an explicit mechanism for controlling the degree of generalization from the labeling functions.
The latter three methods have been developed without any mechanism to control generalization. 

Experiments are conducted for four classifications tasks: sentiment analysis, spam detection, detection of the spouse relationship and question classification. Four data sets that are common in weak supervision are used, which are SPAM, SPOUSE, IMDb and TREC, see Section \ref{sec:datasets}.

\subsection{Models} \label{ref:models}
We use a pre-trained DistilBERT language model to encode the input texts.
Similar to BERT \citep{devlin-etal-2019-bert}, DistilBERT is a masked transformer language model, which is a smaller, lighter, and faster version leveraging knowledge distillation while retaining 97\% of BERT's language understanding capabilities \citep{DistilBERT}.

To arrive with the DistilBERT input encodings we first tokenize the texts using the DistilBERT tokenizer. 
After that, the tokenized input is converted to the DistilBERT transformer encoding, consisting of the input ids as well as the attention mask. 
We use that representation across \KnowMAN{}, \Snorkel{} and \MV{} models.

For \wea{} we encode the input using RoBERTa \citep{liu2019roberta}, a optimized version of the BERT language model, because DistilBERT is not supported by \wea{}.

\subsubsection{\KnowMAN{}}
In previous work we proposed \KnowMAN{} \citep{Mrz2021KnowMANWS}.
The ultimate goal of \KnowMAN{} is to learn a feature representation that is invariant to the labeling functions which annotated the weakly supervised data. 
We showed that this representation is more general and more robust to incorrect classes that have been assigned by the labeling functions. 

\begin{figure}
    \centering
    \includegraphics[width=\textwidth]{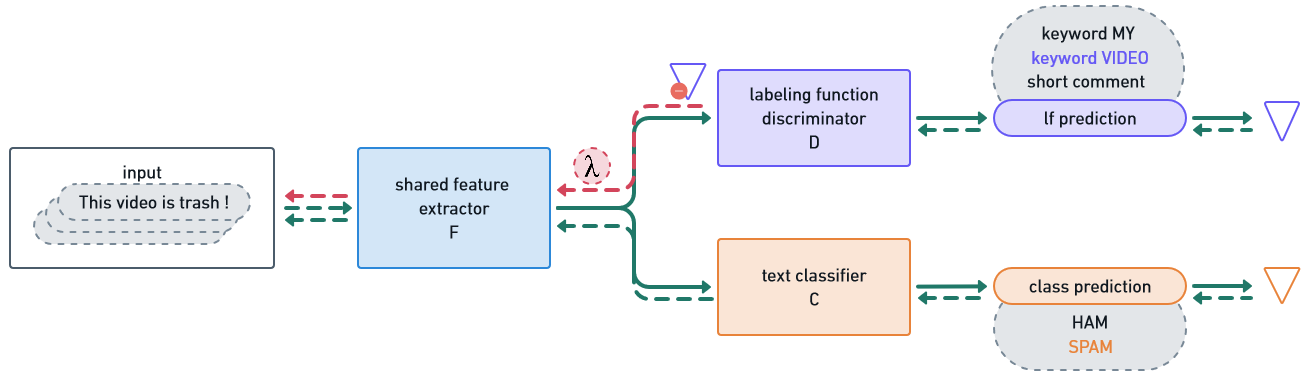}
    \vspace*{1.5cm}
    \caption{\KnowMAN{} architecture. One iteration for a pass of one batch of inputs. The correct class and labeling function for the example instance are highlighted. 
    The parameters of $\mathpzc{C}$ and $\mathpzc{F}$ are updated together, the labeling function discriminator $\mathpzc{D}$ is updated with a separate optimizer. 
    Solid lines indicate forward, dashed lines the backward passes.
    $\nabla$ indicates the (reversed) gradient.
    Before entering $\mathpzc{F}$ the gradient is flipped. 
    The influence of $\mathpzc{D}$ can be controlled by hyperparameter $\lambda$. Strength of $\lambda$ equals generalization strength.}
    
    \label{fig:KnowMAN}
\end{figure}

The architecture is designed as an adversarial model and contains three modules:  i) a shared feature extractor $\mathpzc{F}$, ii) a classifier $\mathpzc{C}$ and iii) a labeling function discriminator $\mathpzc{D}$. See Figure \ref{fig:KnowMAN} for an illustration of the architecture.
The classifier $\mathpzc{C}$ is trained to predict the labels of a downstream task. 
The gradient of the loss function is used to optimize the classifier itself as well as the shared feature extractor.
At the same time, the discriminator $\mathpzc{D}$ is learned to distinguish between the different labeling functions and should predict which of the labeling functions was responsible for labeling an instance. 
The gradient of the discriminators loss function is used to optimize $\mathpzc{D}$. 
In addition, the \emph{reversed} gradient of $\mathpzc{D}$ is used to optimize the feature extractor $\mathpzc{F}$. 
This adversarial update leads to a weakening of the labeling function discrimination information and therefore to a better generalization. 
\KnowMAN{} uses a hyperparameter $\lambda$ to control the level of weakening the signals. 
Consequently, \XPASC{} allows us to study how changes in $\lambda$ affect the degree of generalization of trained \KnowMAN{} models.

\KnowMAN{} is implemented as follows: the discriminator $\mathpzc{D}$ is trained with a separate optimizer than $\mathpzc{C}$ and $\mathpzc{F}$.
When $\mathpzc{D}$ is trained, the parameters of $\mathpzc{C}$ and $\mathpzc{F}$ are frozen and vice versa. 
The losses for both, the classifier and the discriminator,  are computed using negative log-likelihood (NLL). 
The classification NLL can be formalized as:

\begin{equation}
    \mathpzc{L}_C(\hat{c_i}, c_i) = - \log P (\hat{c_i} = c_i)
\end{equation}

where $c_i$ is the (weakly supervised) annotated class and $\hat{c_i}$ is the prediction of the classifier module $\mathpzc{C}$, for a training sample $i$. 
Analogously, the NLL for the labeling function discriminator is defined as: 

\begin{equation}
    \mathpzc{L_D}(\hat{{l}_i}, l_i) = - \log P (\hat{{l}_i} = l_i)
\end{equation}

where $l_i$ is the actual labeling function used for annotating sample $i$ and $\hat{{l}_i}$ is the predicted labeling function by the discriminator $\mathpzc{D}$. 

The results of the experiments with \KnowMAN{} are shown in Table \ref{tab:res}.
As mentioned above, we encode the inputs with DistilBERT. In \cite{Mrz2021KnowMANWS} we did that for the experiments with \KnowMAN{} as well and additionally encoded the input with TF-IDF. Table \ref{tab:res} reports the results for experiments with both, DistilBERT and TF-IDF encodings. 
The baselines are a \MV{} model as well a \Snorkel{} model. 
The functionality of both models is explained in section \ref{ModelsNoGen}.
For DistilBERT encoded input we also trained a fine-tuned DistilBERT model and utilized it for prediction. 

We refer to the \KnowMAN{} model with a $\lambda$ value of zero as \bKM{}. 
Setting $\lambda$ to zero means disabling the generalization mechanism, because the feature extractor $\mathpzc{F}$ is blind for the loss of the discriminator $\mathpzc{D}$.
\KnowMAN{} TF-IDF and \KnowMAN{} DistilBERT refer to a \KnowMAN{} model with optimal $\lambda$ (tuned on the dev set through Bayesian hyperparameter optimization) for the respective dataset.
\KnowMAN{} is able to outperform the baselines for all data sets. The only exception is fine-tuned DistilBERT, which performs better for SPAM.

\begin{table}
\caption{\KnowMAN{} results on the test sets.}
    \centering
    \small
    \begin{tabular}{l|c|ccc|c}
         & SPAM & &  SPOUSE &  & IMDb  \\ 
         \hline
         & Acc & P & R & F1 & Acc  \\ \hline
      \MV{} TF-IDF   & 0.87 &  0.12 & \textbf{0.83} & 0.20 & 0.65 \\
      \bKM{} TF-IDF & 0.91 & 0.12 & 0.76 & 0.21 & 0.75 \\ 
      \Snorkel{} TF-IDF & 0.81 & \textbf{0.18} & 0.63 & 0.28 & 0.50  \\ 
      \KnowMAN{} TF-IDF & \textbf{0.94} & 0.16 & 0.72 & \textbf{0.35} & \textbf{0.77} \\ \hline 
      Fine-tuned DistilBERT & \textbf{0.92} & 0.14 & 0.78 & 0.24 & 0.70 \\
      \MV{} DistilBERT & 0.87 & 0.09 & \textbf{0.90} & 0.17 & 0.67 \\ 
      \bKM{} DistilBERT & 0.86 & 0.18 & 0.80 & 0.29 & 0.74 \\ 
      \Snorkel{} DistilBERT & 0.88 & 0.13 & 0.70 & 0.23 & 0.49 \\
      \KnowMAN{} DistilBERT & 0.90 & \textbf{0.27} & 0.67 & \textbf{0.39} & \textbf{0.76} \\ \hline 
    \end{tabular}
    \label{tab:res}
\end{table}

\subsubsection{Models without generalization mechanism} \label{ModelsNoGen}
We also study three methods with no generalization control. This demonstrates that \XPASC{} allows highlighting generalization across different approaches.
The first method is majority voting. 
The second approach follows the data programming paradigm (DP) proposed by \cite{Ratner2016DataPC}. 
The third approach \citep{weasel}, WeaSEL, aims to learn in an end-to-end fashion from the labeling function output directly.\\ 

\textbf{\MV{}}.
For the majority vote classifier a matrix that holds the mapping between each labeling function and the corresponding class (the labeling function it is associated with) is needed. 
For each instance of the train set it is checked which labeling functions apply. 
Based on this information it is looked up which class each matching labeling function would assign to the instance and if there is a majority for one class among them. 
If so, the class is assigned. 
If not, the class is either chosen at random from among the matching classes, or another special class is assigned to this instance. 
After labeling the training data in this way, we use the uncased DistilBERT model provided by Hugging Face \citep{Wolf2019HuggingFacesTS} as the prediction model. 
\\

\textbf{\Snorkel{}}.
We also compare to training models on labels denoised by \Snorkel{} (Data Programming) \citep{snorkel}.
To do this, we use Knodle's \Snorkel{} wrapper \citep{knodle}, where first a generative \Snorkel{} model is learned, generating weak labels for the instances, and then a classification model (used for prediction) is trained with those labels.
\Snorkel{} works with a set of given labeling functions and learns a label model that focuses on the conflicts and agreements between the labeling functions to estimate their accuracy. For each labeling function an accuracy value is estimated to weigh their votes on each instance. Taking into account the weighted labeling functions, the label model can assign a probabilistic class to each instance and arrives with a weakly supervised data set.
As with \MV{}, the prediction model trained on the weak labels is uncased DistilBERT. The cross-entropy loss is optimzed on the probabilistic \Snorkel{} labels.
\\

\textbf{\wea{}}.
In addition, we train models with \wea{}, an end-to-end model for weak supervision that does not take the noisy weak class labels, but the labeling function output as input for model training \citep{weasel}. 
The approach produces accuracy scores for each labeling source (in our case labeling function) and trains both a neural encoder and a downstream model at the same time on the same loss by using each other's predicted labels as input.
We use the \wea{} implementation of \citet{wrench}, train with the default hyperparameters and use RoBERTa as an encoder.

\begin{table}[t!]
    \centering
    \small
    \caption{Statistics on data set sizes. \textit{Size after filtering} is the size of the data sets after instances without labeling function matches were filtered out.}
    \begin{tabular}{l|c|c|c|c}
    \hline
       Data set  & original size & size after filtering & percentage of original size & labeling functions\\ \hline
         SPAM & 1586  & 1382 & 87.1 & 10 \\
         TREC & 4965 & 4723 & 95.1\% & 68 \\
         SPOUSE & 22254 & 5734 & 25.8\% & 9\\ 
         IMDb & 40000 & 39998 & 99.9\% & 6786\\  \hline 
    \end{tabular}
    
    \label{tab:datasets}
\end{table}

\subsection{Data sets} \label{sec:datasets}
For our experiments we use four standard data sets for weak supervision.
In addition to the three binary data sets covered by \KnowMAN{} (SPAM, SPOUSE, IMDb) we also study one multi class data set (TREC). 
While SPAM, TREC and IMDb are classification tasks, SPOUSE addresses relation extraction.

\subsubsection{SPAM}
\Snorkel{} provides a small subset of the YouTube comments data set \citep{spam} where the task is to classify whether a text is relevant to a certain YouTube video or contains spam. 
Ten different labeling functions are used to assign the classes, mostly based on keywords and regular expressions. 
In contrast to other datasets, no development set is provided for SPAM, which is not relevant for \XPASC{} but for downstream task training.

\subsubsection{TREC}
Another text classification data set is TREC which was proposed by \citep{li-roth-2002-learning} and addresses question classification.
The data set contains automatically retrieved as well as manually constructed questions from six different classes.
Multiple classes can be assigned for each instance, but the authors chose to design TREC as a single-class dataset.
Therefore, the data set was manually annotated with one class per instance.
However, the result of the initial overlapping of classes for each instance is that TREC is difficult to learn.
We use the version of the data set provided by \citet{wrench} within the WRENCH framework, containing 68 keyword-based labeling functions which have been generated by \citet{trecrules}.

\subsubsection{SPOUSE}
This data set addresses a binary relation extraction problem and aims to identify the \emph{spouse} relation in text snippets. It has also been created by \Snorkel{} and is based on the Signal Media One-Million News Articles Data set \citep{Corney2016WhatDA}. 
The nine labeling functions use information from a knowledge base, keywords and patterns. 
One peculiarity of this data set is that it is very skewed, with over 90\% of the instances not holding a spouse relation. 

\subsubsection{IMDb} 
The largest data set we use is IMDb, which contains movie reviews and is based on the data set from \cite{maas-etal-2011-learning}. We use the IMDb version compiled by \cite{knodle}. All of the labeling functions used for this data set are occurrences of positive and negative keywords from \cite{hu}.
The addressed task for IMDb is binary sentiment analysis, classifying the reviews as either positive or negative.
Unlike for the other two data sets, there are $6800$ labeling functions for IMDb, which constitutes a particular challenge to the \Snorkel{} denoising framework.

\section{Experiments with \XPASC{}}
We present here the setup and findings of our analysis of different models, using \XPASC{}. 
We evaluated \XPASC{} both quantitatively across all models and with a qualitative feature analysis for \KnowMAN{}.

\subsection{Evaluation settings}
Since we want to calculate the correlations in \XPASC{} on a representative amount of data for one data set, we use the train sets for the \XPASC{} computations.
Moreover, in a practical weak supervision setting (where \XPASC{} might be used, e.g. for model selection), the existence of labeled development and test sets cannot be assumed, and the \XPASC{} calculation needs to rely on weakly labeled training data only.

When using weak supervision to assign classes with labeling functions it can happen that instances do not have a labeling function match. 
Especially, for SPAM and SPOUSE many instances lack a labeling function match.
Therefore, we filter out those instances with no labeling function matches for all our experiments and arrive with smaller data sets. 
For IMDb and TREC the number of instances does only change slightly, since there are very few instances without labeling function matches. 
Fortunately, the already very small SPAM data set does not get much smaller after filtering.
For SPOUSE, we observe the greatest difference and only 25\% of the original data set remain after filtering. 
See Table \ref{tab:datasets} for the data set sizes of the train sets. 

In contrast to the results reported in \cite{Mrz2021KnowMANWS}, we average results across 15 different seeds for SPAM and SPOUSE and across 5 different seeds for TREC to achieve more stable results. 
Due to the size of IMDb one \XPASC{} run takes three days. 
To consume less resources, we performed one \XPASC{} run with one seed for this data set only.

In our study of \KnowMAN{} with \XPASC{}, we experiment with different values of $\lambda$.
As the hyperparameter $\lambda$ is intended to control the degree of generalization, this 
sheds light onto the the functionality of \KnowMAN{}.  
%gives better insights into the functionality of \KnowMAN{}.  
Specifically, it is useful for examining the hypothesis whether the model is able to generalize from the labeling functions when tuning $\lambda$.
Our expectation is that the higher the value chosen for $\lambda$, the higher the \XPASC{} result.
With the experiments in this paper, we want to confirm that expectation and validate \KnowMAN{} with \XPASC{} and vice versa. 
For \MV{}, \Snorkel{} and \wea{} we do not have a presumption of the \XPASC{} result, but assume that the score could be higher than for \KnowMAN{} because these models have fewer parameters. 

%SPAM, TREC
\begin{figure}[t!]
\centering
\begin{minipage}{.5\textwidth}
 \centering
\begin{tikzpicture}[scale=0.7]
\pgfplotsset{every axis y label/.append style={rotate=270, yshift=3cm, xshift=2cm}}
  \begin{axis}[
    ybar,
    symbolic x coords={mv, snorkel, wea, bkm, km},
    bar width=0.6cm, bar shift=-0.3cm,
    xtick = {mv, snorkel, wea, bkm, km},
    xticklabels={\MV{}, \Snorkel{}, \wea{}, \bKM{}, \KnowMAN{}},
    axis y line*=left,
    ylabel=\XPASC{},
    height=6cm,width=9cm,
    xticklabel style = {font=\small},
    x tick label style={rotate=45, anchor=east, align=left}
    ]
    \addplot[fill = darklavender] %, postaction={pattern=north east lines}] 
    coordinates {
    (mv, 0.999746662) 
    (snorkel, 0.999886124) 
    (wea, 1.000231975)
    (bkm, 0.997752547) 
    (km, 0.999561236)
    };
  \label{generalization}
  \end{axis}
  \pgfplotsset{every axis y label/.append style={xshift=-3.5cm}}
  \begin{axis}[
    legend style={at={(0.68, 1.4)},anchor=north east},
    ybar,
    symbolic x coords={mv, snorkel, wea, bkm, km},
    bar width=0.6cm, bar shift=0.3cm,
    xtick = {mv, snorkel, wea, bkm, km},
    xticklabels={\MV{}, \Snorkel{}, \wea{}, \bKM{}, \KnowMAN{}},
    axis y line*=right,
    ylabel=accuracy,
    height=6cm,width=9cm,
    hide x axis,
    ]
    \addlegendimage{/pgfplots/refstyle=generalization}\addlegendentry{generalization}
    \addplot[fill=lightcornflowerblue] %,  postaction={pattern=horizontal lines}] 
    coordinates {
        (mv, 0.86) 
        (snorkel, 0.88) 
        (wea, 0.93)
        (bkm, 0.88) 
        (km, 0.90)
      };
  \addlegendentry{performance}
  \end{axis}
\end{tikzpicture}
\vspace*{1.5cm}
\caption{Evaluation of \XPASC{} across baselines and \KnowMAN{} models for SPAM. \bKM{} is a \KnowMAN{} model with $\lambda$ set to $0$. \KnowMAN{} for SPAM means a model trained with  $\lambda$ set to $2$.}
\label{fig:SpamModels}
\end{minipage}%
\begin{minipage}{.5\textwidth}
 \centering
\begin{tikzpicture}[scale=0.7]
\pgfplotsset{every axis y label/.append style={rotate=270, yshift=3cm, xshift=2cm}}
  \begin{axis}[
    ybar,
    symbolic x coords={mv, snorkel, wea, bkm, km},
    bar width=0.6cm, bar shift=-0.3cm,
    xtick = {mv, snorkel, wea, bkm, km},
    xticklabels={\MV{}, \Snorkel{}, \wea{}, \bKM{}, \KnowMAN{}},
    axis y line*=left,
    ylabel=\XPASC{},
    height=6cm,width=9cm,
    xticklabel style = {font=\small},
    x tick label style={rotate=45, anchor=east, align=left}
    ]
    \addplot[fill = darklavender] %, postaction={pattern=north east lines}] 
    coordinates {
    (mv, 0.995604392) 
    (snorkel, 0.999641471) 
    (wea, 0.998865432)
    (bkm, 0.987416541) 
    (km, 0.986878307)
    };
  \label{generalization}
  \end{axis}
  \pgfplotsset{every axis y label/.append style={xshift=-3.5cm}}
  \begin{axis}[
    legend style={at={(0.68, 1.4)},anchor=north east},
    ybar,
    symbolic x coords={mv, snorkel, wea, bkm, km},
    bar width=0.6cm, bar shift=0.3cm,
    xtick = {mv, snorkel, wea, bkm, km},
    xticklabels={\MV{}, \Snorkel{}, \wea{}, \bKM{}, \KnowMAN{}},
    axis y line*=right,
    ylabel=accuracy,
    height=6cm,width=9cm,
    hide x axis,
    ]
    \addlegendimage{/pgfplots/refstyle=generalization}\addlegendentry{generalization}
    \addplot[fill=bananayellow] %,  postaction={pattern=horizontal lines}] 
    coordinates {
        (mv, 0.58) 
        (snorkel, 0.46) 
        (wea, 0.57)
        (bkm, 0.58) 
        (km, 0.63)
      };
  \addlegendentry{performance}
  \end{axis}
\end{tikzpicture}
\vspace*{1.5cm}
\caption{Evaluation of \XPASC{} across baselines and \KnowMAN{} models for TREC. \bKM{} is a \KnowMAN{} model with $\lambda$ set to $0$. \KnowMAN{} for TREC means a model trained with  $\lambda$ set to $0.001$.}
\label{fig:TRECModels}
\end{minipage}%
\end{figure}
%Across models Spouse & IMDB
\begin{figure}[t!]
\centering
\begin{minipage}{.5\textwidth}
   \centering
\begin{tikzpicture}[scale=0.7]
\pgfplotsset{every axis y label/.append style={rotate=270, yshift=3cm, xshift=2cm}}
  \begin{axis}[
    ybar,
    symbolic x coords={mv, snorkel, wea, bkm, km},
    bar width=0.6cm, bar shift=-0.3cm,
    xtick = {mv, snorkel, wea, bkm, km},
    xticklabels={\MV{}, \Snorkel{}, \wea{}, \bKM{}, \KnowMAN{}},
    axis y line*=left,
    ylabel=\XPASC{},
    height=6cm,width=9cm,
    xticklabel style = {font=\small},
    x tick label style={rotate=45, anchor=east, align=left}
    ]
    \addplot[fill = darklavender] %, postaction={pattern=north east lines}] 
    coordinates {
    (mv, 0.999675074) 
    (snorkel, 0.99996464)
    (wea, 0.998909424)
    (bkm, 0.999014062) 
    (km, 0.999021169)
        };
  \label{generalization}
  \end{axis}
  \pgfplotsset{every axis y label/.append style={xshift=-3.5cm}}
  \begin{axis}[
    legend style={at={(0.68, 1.4)},anchor=north east},
    ybar,
    symbolic x coords={mv, snorkel, wea, bkm, km},
    bar width=0.6cm, bar shift=0.3cm,
    xtick={mv, snorkel, wea, bkm, km},
    xticklabels={\MV{}, \Snorkel{}, \wea{}, \bKM{}, \KnowMAN{}},
    axis y line*=right,
    ylabel=F1,
    ymin = 0., ymax = 25,
    height=6cm,width=9cm,
    hide x axis,
    ]
    \addlegendimage{/pgfplots/refstyle=generalization}\addlegendentry{generalization}
    \addplot[fill=cadmiumorange] %, postaction={pattern=vertical lines}] 
    coordinates {
        (mv, 16.57) 
        (snorkel, 15.26) 
        (wea, 0.2)
        (bkm, 20.55) 
        (km, 22.28)
      };
  \addlegendentry{performance}
  \end{axis}
\end{tikzpicture}
\vspace*{1.5cm}
\caption{Evaluation of \XPASC{} across baselines and \KnowMAN{} models for SPOUSE. \bKM{} means a \KnowMAN{} model with $\lambda$ set to $0$. \KnowMAN{} for SPOUSE means a model trained with  $\lambda$ set to $0.5$.}
\label{fig:SpouseModels}
\end{minipage}%
\begin{minipage}{.5\textwidth}
   \centering
\begin{tikzpicture}[scale=0.7]
\pgfplotsset{every axis y label/.append style={rotate=270, yshift=3cm, xshift=2cm}}
  \begin{axis}[
    ybar,
    symbolic x coords={mv, snorkel, wea, bkm, km},
    bar width=0.6cm, bar shift=-0.3cm,
    xtick={mv, snorkel, wea, bkm, km},
    xticklabels={\MV{}, \Snorkel{}, \wea{}, \bKM{}, \KnowMAN{}},
    axis y line*=left,
    ylabel=\XPASC{},
    height=6cm,width=9cm,
    xticklabel style = {font=\small},
    x tick label style={rotate=45, anchor=east, align=left}
    ]
    \addplot[fill = darklavender] %, postaction={pattern=north east lines}] 
    coordinates {
    (mv, 0.999850705) 
    (snorkel, 1.) 
    (wea, 0.9999976201140742)
    (bkm, 0.999703447) 
    (km, 0.999732707)
        };
  \label{generalization}
  \end{axis}
  \pgfplotsset{every axis y label/.append style={xshift=-3.5cm}}
  \begin{axis}[
    legend style={at={(0.68, 1.4)},anchor=north east},
    ybar,
    symbolic x coords={mv, snorkel, wea, bkm, km},
    bar width=0.6cm, bar shift=0.3cm,
    xtick={mv, snorkel, wea, bkm, km},
    xticklabels={\MV{}, \Snorkel{}, \wea{}, \bKM{}, \KnowMAN{}},
    axis y line*=right,
    ylabel=accuracy,
    height=6cm,width=9cm,
    hide x axis,
    ]
    \addlegendimage{/pgfplots/refstyle=generalization}\addlegendentry{generalization}
    \addplot[fill=asparagus] %, postaction={pattern=grid}] 
    coordinates {
        (mv, 67.36) 
        (snorkel, 50.) 
        (wea, 76.7)
        (bkm, 73.7) 
        (km, 75.32)
      };
  \addlegendentry{performance}
  \end{axis}
\end{tikzpicture}
\vspace*{1.5cm}
\caption{Evaluation of \XPASC{} across baselines and \KnowMAN{} models for IMDb. \bKM{} means a \KnowMAN{} model with $\lambda$ set to $0$. \KnowMAN{} for IMDb means a model trained with  $\lambda$ set to $0.5$.}
\label{fig:IMDbModels}
\end{minipage}%
\end{figure}

\subsection{Quantitative evaluation}
Our quantitative evaluation includes \XPASC{} results across all models mentioned in section \ref{ref:models}, as well as the results for different $\lambda$ values for \KnowMAN{}. 
In addition, we evaluated \KnowMAN{} with \XPASC{} for both association measures, \chis{} and \PPMI{}.

\subsubsection{\XPASC{} results across all models}
We calculated \XPASC{} with \chis{}-based association for \MV{}, \Snorkel{}, \wea{} and \KnowMAN{} on all data sets. 

The evaluation of SPAM (see Figure \ref{fig:SpamModels}) shows the highest generalization for \wea{} and the \Snorkel{} model achieves the second highest \XPASC{}. 
Generalization scores for \MV{} and the \KnowMAN{} model with the optimal $\lambda$ value of 2 are similar and slightly worse than for \Snorkel{}. 
The generalization of the \bKM{} model with the disabled generalization mechanism is in contrast the lowest. 
The observations are different for the class prediction performance of the models. Here \wea{} and \KnowMAN{} give the highest, whereas \MV{} gives the worst classification accuracy.  

For TREC we observe the highest generalization for \Snorkel{} and \wea{} (see Figure \ref{fig:TRECModels}). 
Both achieve low classification numbers, although their performance differs significantly. 
\Snorkel{} achieves the lowest results and \wea{} achieves results similar to \bKM{} and \MV{}.
\MV{} shows a similar classification performance as \KnowMAN{} but much higher generalization. 
The \KnowMAN{} models achieve the best classification performance, but have the lowest \XPASC{}.
Unlike for the other data sets, using \KnowMAN{} decreases the \XPASC{} value slightly.

The evaluation of SPOUSE (see Figure \ref{fig:SpouseModels}) shows a picture similar to SPAM. 
Again, \Snorkel{} achieves the highest \XPASC{}. 
The scores of both \KnowMAN{} models are close to the generalization score of the \wea{} model. 
The difference between \bKM{} and \KnowMAN{} is smaller than in the other data sets. 
In terms of performance, the models are ranked differently. 
Both \KnowMAN{} models perform better than \Snorkel{} or \MV{}. 
Indeed, the performance of \Snorkel{} is the lowest, while this model has a high generalization value.
The \wea{} model gives unreasonably low results for both, \XPASC{} and F1 score, and like \citet{sepll} we assume that this is due to the fact that they did not integrate large pre-trained language models like RoBERTa in their original work. 
 
For IMDb the results are in agreement with the insights of the other data sets (see Figure \ref{fig:IMDbModels}).
Again, \Snorkel{} gives the highest \XPASC{} and \wea{} the second highest \XPASC{} result. 
Since \bKM{} and \KnowMAN{} are only different by a $\lambda$ value of 0.5, their generalization scores are close to each other. 
Again, \MV{} reaches a slightly higher \XPASC{} than the \KnowMAN{} models. 
The classification accuracy gives the same ranking as for SPOUSE, except for the \wea{} model, which performs best on the IMDb data set.
Both \KnowMAN{} models perform better than \Snorkel{} and \MV{}, while having smaller \XPASC{} results.

Overall we observe that models without explicit generalization modeling achieve higher \XPASC{} values than \KnowMAN{}. 
This can be explained by the fact that these models employ a smaller number of layers, thus are less complex and enable greater generalization more easily. 
The more complex a model becomes and the more parameters it consists of, the more likely it is to overfit and thus generalization is more difficult to achieve. 
Another observation that can easily be drawn from the plots is that performance and generalization are not related one-to-one. 
Indeed, it seems like greater generalization often hinders performance. 
An explanation for this correlation may be that more generalization leads to less (over-)fitting, what can also harm performance. 
The works of \citet{benignoverfitting} and \cite{zhang2020overfitting} show that models that overfit to noisy data still can achieve good performance and that the noise doesn't harm the performance as much as expected. 
Still, \citet{sanyal2020benign} claim that overfitting might not harm the performance of a model, but its robustness and that too much overfitting makes a model vulnerable (e.g. to adversarial attacks).

% Spam plots
\begin{figure}[t!]
\centering
\begin{minipage}{.5\textwidth}
    \centering
    \begin{tikzpicture}[scale=0.7]
    \pgfplotsset{every axis y label/.append style={rotate=270, yshift=3.5cm, xshift=2cm}}
  \begin{axis}[
    symbolic x coords={0, 0.5, 1, 2, 4},
    xtick={0, 0.5, 1, 2, 4},
    axis y line* = left, % the '*' avoids arrow heads
    xlabel = lambda $\lambda$,
    ylabel = \XPASC{}]
    \addplot[darklavender, domain=700:2000, style={ultra thick}]
    coordinates {
    (0, 0.997752547)
    (0.5, 0.998066513)
    (1, 0.998740764)
    (2, 0.999561236)
    (4,0.999878733)
    };
  \end{axis}
  \pgfplotsset{every axis y label/.append style={xshift=-3.5cm}}
  \begin{axis}[
    legend style={at={(0.72,1.3)},anchor=north east},
    symbolic x coords={0, 0.5, 1, 2, 4},
    xtick={0, 0.5, 1, 2, 4},
    ymin = 86.5, ymax = 90.5,
    hide x axis,
    axis y line*=right,
    ylabel=accuracy]
    \addlegendimage{/pgfplots/refstyle=generalization}\addlegendentry{generalization}
    \addplot[lightcornflowerblue,domain=750:1800, style={ultra thick}]
    coordinates {
    (0,87.57333333)(0.5, 87.84)(1,88.46)(2,90.05333333)(4,87.22666667)
    };
    \addlegendentry{performance}
  \end{axis}
\end{tikzpicture}
    \vspace*{0.8cm}
    \caption{Chi-square-based \XPASC{} and accuracy for SPAM across different $\lambda$ values.}
    \label{fig:SpamChi2}
\end{minipage}%
\begin{minipage}{.5\textwidth}
   \centering
        \begin{tikzpicture}[scale=0.7]
    \pgfplotsset{every axis y label/.append style={rotate=270, yshift=3.5cm, xshift=2cm}}
  \begin{axis}[
    symbolic x coords={0, 0.5, 1, 2, 4},
    xtick={0, 0.5, 1, 2, 4},
    axis y line* = left, % the '*' avoids arrow heads
    ylabel=generalization,
    xlabel = lambda $\lambda$,
    ylabel = \XPASC{}]
    \addplot[darklavender, domain=700:2000, style={ultra thick}]
    coordinates {
    (0, 0.99533147)
    (0.5, 0.995861212)
    (1, 0.997235618)
    (2, 0.9988804)
    (4, 0.999650685)
    };
  \end{axis}
  \pgfplotsset{every axis y label/.append style={xshift=-3.5cm}}
  \begin{axis}[
    legend style={at={(0.72,1.3)},anchor=north east},
    symbolic x coords={0, 0.5, 1, 2, 4},
    xtick={0, 0.5, 1, 2, 4},
    ymin = 86.5, ymax = 90.5,
    hide x axis,
    axis y line*=right,
    ylabel=accuracy]
    \addlegendimage{/pgfplots/refstyle=generalization}\addlegendentry{generalization}
    \addplot[lightcornflowerblue,domain=750:1800, style={ultra thick}]
    coordinates {
    (0,87.57333333)(0.5, 87.84)(1,88.46)(2,90.05333333)(4,87.22666667)
    };
    \addlegendentry{performance}
  \end{axis}
\end{tikzpicture}
    \vspace*{0.8cm}
    \caption{PPMI-based \XPASC{} and accuracy for SPAM across different $\lambda$ values.}
    \label{fig:SpamPPMI}
\end{minipage}
\end{figure}

% TREC plots
\begin{figure}[t!]
\centering
\begin{minipage}{.5\textwidth}
    \centering
    \begin{tikzpicture}[scale=0.7]
    \pgfplotsset{every axis y label/.append style={rotate=270, yshift=3.5cm, xshift=2cm}}
  \begin{semilogxaxis}[
    xticklabels={$10^{-6}$, $\textcolor{white}{1}0^{\textcolor{white}{-5}}$, $10^{-4}$, $10^{-3}$, $10^{-2}$, $10^{-1}$, $1$},
    axis y line* = left, % the '*' avoids arrow heads
    ylabel=\XPASC{},
    xlabel = lambda $\lambda$,
    ylabel = XPASC]
    \addplot[darklavender, domain=700:2000, style={ultra thick}]
    coordinates {
    (0.00001, 0.987416541)
    (0.0001, 0.987286129)
    (0.001, 0.986878307)
    (0.01, 0.989066936)
    (0.1, 0.987418182)
    (1, 0.996592004)
    };
  \end{semilogxaxis}
  \pgfplotsset{every axis y label/.append style={xshift=-3.5cm}}
  \begin{semilogxaxis}[
    legend style={at={(0.72,1.3)},anchor=north east},
    hide x axis,
    axis y line*=right,
    ylabel=accuracy]
    \addlegendimage{/pgfplots/refstyle=generalization}\addlegendentry{generalization}
    \addplot[bananayellow,domain=750:1800, style={ultra thick}]
    coordinates {
    (0.00001, 57.5)(0.0001, 60.9)(0.001, 62.7)(0.01, 58.6)(0.1, 59.7)(1, 59.9)
    };
    \addlegendentry{performance}
  \end{semilogxaxis}
\end{tikzpicture}
    \vspace*{0.8cm}
    \caption{Chi-squared-based \XPASC{} and accuracy for TREC across different $\lambda$ values.}
    \label{fig:TRECChi2}
\end{minipage}%
\begin{minipage}{.5\textwidth}
    \centering
    \begin{tikzpicture}[scale=0.7]
    \pgfplotsset{every axis y label/.append style={rotate=270, yshift=3.5cm, xshift=2cm}}
  \begin{semilogxaxis}[
    xticklabels={$10^{-6}$, $\textcolor{white}{1}0^{\textcolor{white}{-5}}$, $10^{-4}$, $10^{-3}$, $10^{-2}$, $10^{-1}$, $1$},    
    axis y line* = left, % the '*' avoids arrow heads
    ylabel=\XPASC{},
    xlabel = lambda $\lambda$,
    ylabel = XPASC]
    \addplot[darklavender, domain=700:2000, style={ultra thick}]
    coordinates {
    (0.00001, 0.985610116)
    (0.0001, 0.982900344)
    (0.001, 0.982440655)
    (0.01, 0.985274889)
    (0.1, 0.984845321)
    (1, 0.994123934)
    };
  \end{semilogxaxis}
  \pgfplotsset{every axis y label/.append style={xshift=-3.5cm}}
  \begin{semilogxaxis}[
    legend style={at={(0.72,1.3)},anchor=north east},
    hide x axis,
    axis y line*=right,
    ylabel=accuracy]
    \addlegendimage{/pgfplots/refstyle=generalization}\addlegendentry{generalization}
    \addplot[bananayellow,domain=750:1800, style={ultra thick}]
    coordinates {
    (0.00001, 57.5)(0.0001, 60.9)(0.001, 62.7)(0.01, 58.6)(0.1, 59.7)(1, 59.9)
    };
    \addlegendentry{performance}
  \end{semilogxaxis}
\end{tikzpicture}
    \vspace*{0.8cm}
    \caption{PPMI-based \XPASC{} and accuracy for TREC across different $\lambda$ values.}
    \label{fig:TRECPPMI}
\end{minipage}
\end{figure}

% Spouse plots
\begin{figure}
\centering
\begin{minipage}{.5\textwidth}
    \centering
    \begin{tikzpicture}[scale=0.7]
    \pgfplotsset{every axis y label/.append style={rotate=270, yshift=3.5cm, xshift=2cm}}
  \begin{axis}[
    symbolic x coords={0, 0.5, 1, 2, 4},
    xtick={0, 0.5, 1, 2, 4},
    axis y line* = left, % the '*' avoids arrow heads
    ylabel=\XPASC{},
    xlabel = lambda $\lambda$,
    ylabel = XPASC]
    \addplot[darklavender, domain=700:2000, style={ultra thick}]
    coordinates {
    (0, 0.999014062)
    (0.5, 0.999021169)
    (1, 0.999176572)
    (2, 0.999719968)
    (4, 0.99993856)
    };
  \end{axis}
  \pgfplotsset{every axis y label/.append style={xshift=-3.5cm}}
  \begin{axis}[
    legend style={at={(0.72,1.3)},anchor=north east},
    symbolic x coords={0, 0.5, 1, 2, 4},
    xtick={0, 0.5, 1, 2, 4},
    ymin = 12, ymax = 24,
    hide x axis,
    axis y line*=right,
    ylabel=F1]
    \addlegendimage{/pgfplots/refstyle=generalization}\addlegendentry{generalization}
    \addplot[cadmiumorange,domain=750:1800, style={ultra thick}]
    coordinates {
    (0, 20.55)(0.5, 22.28)(1, 22.25)(2, 19.92)(4, 14.98)
    };
    \addlegendentry{performance}
  \end{axis}
\end{tikzpicture}
    \vspace*{0.8cm}
    \caption{Chi-squared-based \XPASC{} and F1 for SPOUSE across different $\lambda$ values.}
    \label{fig:SpouseChi2}
\end{minipage}%
\begin{minipage}{.5\textwidth}
    \centering
    \begin{tikzpicture}[scale=0.7]
    \pgfplotsset{every axis y label/.append style={rotate=270, yshift=3.5cm, xshift=2cm}}
  \begin{axis}[
    symbolic x coords={0, 0.5, 1, 2, 4},
    xtick={0, 0.5, 1, 2, 4},
    axis y line* = left, % the '*' avoids arrow heads
    ylabel=\XPASC{},
    xlabel = lambda $\lambda$,
    ylabel = XPASC]
    \addplot[darklavender, domain=700:2000, style={ultra thick}]
    coordinates {
    (0, 0.999956911)
    (0.5, 0.999954567)
    (1, 0.999957348)
    (2, 0.999982153)
    (4, 0.999997302)
    };
  \end{axis}
  \pgfplotsset{every axis y label/.append style={xshift=-3.5cm}}
  \begin{axis}[
    legend style={at={(0.72,1.3)},anchor=north east},
    symbolic x coords={0, 0.5, 1, 2, 4},
    xtick={0, 0.5, 1, 2, 4},
    ymin = 12, ymax = 24,
    hide x axis,
    axis y line*=right,
    ylabel=F1]
    \addlegendimage{/pgfplots/refstyle=generalization}\addlegendentry{generalization}
    \addplot[cadmiumorange,domain=750:1800, style={ultra thick}]
    coordinates {
    (0, 20.55)(0.5, 22.28)(1, 22.25)(2, 19.92)(4, 14.98)
    };
    \addlegendentry{performance}
  \end{axis}
\end{tikzpicture}
    \vspace*{0.8cm}
    \caption{PPMI-based \XPASC{} and accuracy for SPOUSE across different $\lambda$ values.}
    \label{fig:SpousePPMI}
\end{minipage}
\end{figure}

%Chi2 IMDb
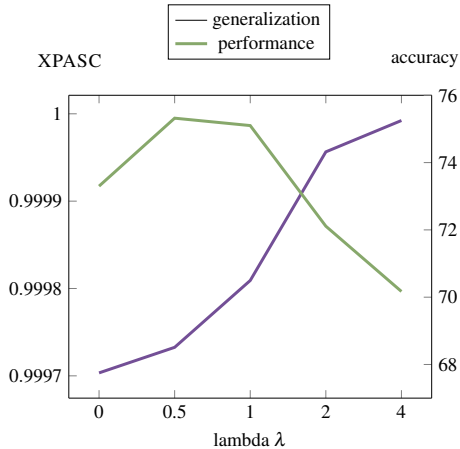
\begin{figure}
\centering
\begin{tikzpicture}[scale=0.7]
\pgfplotsset{every axis y label/.append style={rotate=270, yshift=3.5cm, xshift=2cm}}
  \begin{axis}[
    symbolic x coords={0, 0.5, 1, 2, 4},
    xtick={0, 0.5, 1, 2, 4},
    axis y line* = left, % the '*' avoids arrow heads
    ylabel=generalization,
    xlabel = lambda $\lambda$,
    ylabel = \XPASC{}]
    \addplot[darklavender, domain=700:2000, style={ultra thick}]
    coordinates {
    (0, 0.999703447)
    (0.5, 0.999732707)
    (1,0.999809234)
    (2, 0.9999565)
    (4, 0.99999238)
    };
  \end{axis}
    \pgfplotsset{every axis y label/.append style={xshift=-3.5cm}}
  \begin{axis}[
    legend style={at={(0.72,1.3)},anchor=north east},
    symbolic x coords={0, 0.5, 1, 2, 4},
    xtick={0, 0.5, 1, 2, 4},
    ymin = 67, ymax = 76,
    hide x axis,
    axis y line*=right,
    ylabel=accuracy]
    \addlegendimage{/pgfplots/refstyle=generalization}\addlegendentry{generalization}
    \addplot[asparagus,domain=750:1800, style={ultra thick}]
    coordinates {
    (0, 73.3)(0.5, 75.32)(1, 75.1)(2, 72.11)(4, 70.17)
    };
    \addlegendentry{performance}
  \end{axis}
\end{tikzpicture}
    \vspace*{0.5cm}
    \caption{Chi-squared-based \XPASC{} and accuracy for IMDb across different $\lambda$ values.}
    \label{fig:IMDbChi2}
\end{figure}

\subsubsection{\XPASC{} results across \KnowMAN{} models}
To figure out if the degree of generalization can be controlled via the hyperparameter $\lambda$ in the \KnowMAN{} models, we calculate \XPASC{} for different $\lambda$ values.
We calculate both, \XPASC{} \chis{} and \XPASC{} \PPMI{} for SPAM, TREC and SPOUSE. 
For IMDb we calculate \XPASC{} \chis{} only, to save resources.

Figures \ref{fig:SpamChi2} and \ref{fig:SpamPPMI} show the results for SPAM.
It can be clearly seen that \XPASC{} increases with increasing $\lambda$. 
The performance has its peak when using a $\lambda$ value of 2.0.
Both \XPASC{} curves of the two association options are very similar in their shape, though, we find slightly smaller values for \XPASC{} \chis{}.
To compare the two options for calculating association we draw scatter plots that depict the performance and generalization per run and seed (see Figures \ref{fig:spamscatterchi2} and \ref{fig:spamscatterppmi}). 
For the \chis{} option the accuracy and \XPASC{} scores are clustered clearly recognizable for each $\lambda$ value.
For \PPMI{} we observe slightly different results. 
Especially for lower $\lambda$s the distributions are more mixed up. 
Still, both plots reflect the clustering of \XPASC{}-values according to the chosen $\lambda$-values, and the performance trends, nicely.

The results for TREC (see Figures \ref{fig:TRECChi2}, \ref{fig:TRECPPMI}) show that  
\KnowMAN{} is sensitive to small values of the hyperparameter $\lambda$ in the multi-class setting.
Using smaller $\lambda$ values (in comparison to binary models) does improve the performance of the TREC model, whereas using greater $\lambda$ values is less effective.
The model with the best classification performance is trained with $\lambda=0.001$.
With regard to \XPASC{}, one can clearly see from the curve that smaller $\lambda$ values increase the generalization to a lesser extent than larger $\lambda$ values.
Moreover, the trends are less clear and effects are more brittle in this setting.

For SPOUSE, we observe a clear positive correlation of \XPASC{} in relation to larger $\lambda$ as well (see Figures \ref{fig:SpouseChi2}, \ref{fig:SpousePPMI}). 
Unlike for the SPAM and IMDb the curve is not strictly monotonically increasing, however. 
The scatter plots with the distributions of all results across the 15 runs show that for those $\lambda$ values that cause the dips in the curve, there are some outliers that cause the lowerings (see Figures \ref{fig:spousescatterchi2} and \ref{fig:spousescatterppmi}). 
In general SPOUSE appears more challenging and unstable, having more outliers for both axes, \XPASC{} and F1 score.
In terms of performance, the optimum is reached with a $\lambda$ value of 0.5 and decreases drastically with $\lambda$ values being higher than 1.

The \XPASC{} curve for IMDb increases monotonically in relation to $\lambda$ (see Figure \ref{fig:IMDbChi2}).
There are no lowerings or peaks in the \XPASC{} curve for this data set. 
As for SPOUSE, the best performance is reached with $\lambda = 0.5$ and decreases with $\lambda$ being higher than 1.

Overall, we can conclude that it is possible to control the degree of generalization for the \KnowMAN{} models by using the respective hyperparameter $\lambda$. 
In addition, the results confirm our assumption that \XPASC{} can reflect the generalization of models.
As for the evaluation across all models, it shows that performance and generalization are not the same. 
The highest performance is not achieved with the greatest degree of generalization.

%Scatter spam
\begin{figure}
\centering
\begin{minipage}{.5\textwidth}
  \centering
    \begin{tikzpicture}[scale=0.65]
          \pgfplotsset{scale only axis}
          \begin{axis}[
            xticklabels={1-(8),1-(8), 1-(6), 1-(4), 1-(2), 1},
            xlabel=\XPASC{},
            ylabel=accuracy,
            legend entries={{$\lambda=0.0$}, {$\lambda=0.5$}, {$\lambda=1.0$}, {$\lambda=2.0$}, {$\lambda=4.0$}},
            legend columns=5,
            legend style={at={(0.98, 1.15)},anchor=north east},
            legend style={/tikz/every even column/.append style={column sep=0.3cm}}],
            ymin = 78, ymax = 94,
          ]
            \addplot[only marks, mark=asterisk, color=blue]
            coordinates{ % plot 0 data set
              (-0.00077643,	    86)   
              (-0.000591465,    87.6)  
              (-0.000549923,    86)
              (-0.000635931,    85.2)
              (-0.000657075,    89.2)
              (-0.000432483,    91.2)
              (-0.000813277,    87.6)
              (-0.000601968,    86)
              (-0.000614098,    90)
              (-0.000629719,    90.4)
              (-0.00066138,     87.2)
              (-0.000625843,    88)
              (-0.000612872,    84.4)
              (-0.000492328,    88)
              (-0.000840937,    86.8)
            }; \label{plot_one}
            \addplot[only marks, mark=triangle, color=orange]
            coordinates{ % plot 05 data set
              (-0.000698687,    86.4)
              (-0.000500917,    87.2)
              (-0.00047420,     84.8)
              (-0.000498023,    87.2)
              (-0.000560134,    90)
              (-0.000362416,    91.6)
              (-0.000505191,    90)
              (-0.000509086,    89.2)
              (-0.000499741,    89.2)
              (-0.000560765,    88.8)
              (-0.000430352,    90.4)
              (-0.000462835,    85.6)
              (-0.000500781,    85.2)
              (-0.000567296,    88)
              (-0.000578167,    84)

            };
            \addplot[only marks, mark=diamond, color=green]
            coordinates{ % plot 1 data set
              (-0.000445865,    85.6)
              (-0.000377324,    86)
              (-0.000309691,    86.8)
              (-0.000290858,    90)
              (-0.000355848,    89.6)
              (-0.000264471,    88.8)
              (-0.000318033,    90)
              (-0.0003661,      90)
              (-0.000379205,    92.4)
              (-0.000396474,    89.6)
              (-0.000308644,    89.2)
              (-0.000309563,    86.9)
              (-0.000334466,    88.4)
              (-0.000215895,    88.4)
              (-0.000418465,    85.2)

            };
             \addplot[only marks, mark=|, color=cyan, style={ultra thick}]
            coordinates{ % plot 2 data set
              (-0.000209967,    89.2)
              (-0.000188653,    91.2)
              (-8.1231E-05,      90.8)
              (-9.15415E-05,    90)
              (-0.000151676,    90.4)
              (-8.2314E-05,     92)
              (-0.000135901,    91.6)
              (-0.000115564,    90)
              (-0.000152778,    90.4)
              (-8.93287E-05,    86.8)
              (-6.8579E-05,     90.4)
              (-0.000124832,    88.8)
              (-0.000120345,    90.4)
              (-0.000104936,    91.2)
              (-0.000166704,    87.6)
            };
            \addplot[only marks, mark=o, color=yellow, style={ultra thick}]
            coordinates{ % plot 4 data set
              (-4.29527E-05,    79.2)
              (-3.6593E-05,     88.8)
              (-2.02901E-05,    88.8)
              (-2.33453E-05,    88)
              (-3.61952E-05,    89.2)
              (-4.49353E-05,    88.8)
              (-2.57412E-05,    86.8)
              (-3.25776E-05,    90)
              (-3.67325E-05,    87.6)
              (-3.71847E-05,    83.6)
              (-7.3909E-05,     88)
              (-2.73598E-05,    87.6)
              (-2.65324E-05,    83.6)
              (-2.04182E-05,    88.8)
              (-3.99051E-05,    89.6)
            };
          \end{axis}
    \end{tikzpicture}
    \vspace*{1.5cm}
    \caption{\XPASC{}-\chis{} and accuracy results of runs across 15 seeds for SPAM. Different colors indicate different $\lambda$ values. 
    Numbers in brackets must be multiplied by the subscript value, $10^{-4}$.}
    \label{fig:spamscatterchi2}
\end{minipage}%
\begin{minipage}{.5\textwidth}
      \centering
    \begin{tikzpicture}[scale=0.65]
          \pgfplotsset{scale only axis,}
          \begin{axis}[
            xticklabels={1-(8),1-(8), 1-(6), 1-(4), 1-(2), 1},
            xlabel=\XPASC{},
            ylabel=accuracy,
            legend entries={{$\lambda=0.0$}, {$\lambda=0.5$}, {$\lambda=1.0$}, {$\lambda=2.0$}, {$\lambda=4.0$}},
            legend columns=5,
            legend style={at={(0.98, 1.15)},anchor=north east},
            legend style={/tikz/every even column/.append style={column sep=0.3cm}}],
            xmin = -0.009, xmax = 0,
            ymin = 78, ymax = 94,
          ]
            \addplot[only marks, mark=asterisk, color=blue]
            coordinates{ % plot 0 data set
              (-0.005220784,	86)
              (-0.004642392,	87.6)
              (-0.004641062,	86)
              (-0.005232548,	85.2)
              (-0.003878104,	89.2)
              (-0.003329499,	91.2)
              (-0.004268032,	87.6)
              (-0.003430579,	86)
              (-0.003458742,	90)
              (-0.004195495,	90.4)
              (-0.00513026,	    87.2)
              (-0.00819222, 	88)
              (-0.004108009,	84.4)
              (-0.00417888,	    88)
              (-0.006121342,	86.8)
            }; \label{plot_one}
            \addplot[only marks, mark=triangle, color=orange]
            coordinates{ % plot 05 data set
              (-0.005214177,	86.4)
              (-0.00366206,	    87.2)
              (-3.82E-03,   	84.8)
              (-0.004331401,	87.2)
              (-0.004452335,	90)
              (-3.39E-03,   	91.6)
              (-0.003891316,	90)
              (-0.003322862,	89.2)
              (-0.003844368,	89.2)
              (-0.00421865,	    88.8)
              (-0.003924038,	90.4)
              (-0.004413746,	85.6)
              (-0.004248657,	85.2)
              (-0.003501469,	88)
              (-0.005852747,	84)
            };
            \addplot[only marks, mark=diamond, color=green]
            coordinates{ % plot 1 data set
              (-0.003646425,	85.6)
              (-0.002531294,	86)
              (-2.44E-03,	    86.8)
              (-0.002585895,	90)
              (-0.002795896,	89.6)
              (-0.002637784,	88.8)
              (-0.002504415,	90)
              (-0.003020259,	90)
              (-0.002844063,	92.4)
              (-0.002816316,	89.6)
              (-0.002628471,	89.2)
              (-0.00264037,	    86.9)
              (-0.002540345,	88.4)
              (-0.002449937,	88.4)
              (-0.003381844,	85.2)
            };
             \addplot[only marks, mark=|, color=cyan, style={ultra thick}]
            coordinates{ % plot 2 data set
              (-0.002457083,	89.2)
              (-0.00181126,	    91.2)
              (-6.93E-04,	    90.8)
              (-0.000720294,	90)
              (-0.001083946,	90.4)
              (-0.000582967,	92)
              (-0.001509449,	91.6)
              (-0.000972874,	90)
              (-0.001035878,	90.4)
              (-0.000749332,	86.8)
              (-0.000549777,	90.4)
              (-0.000984021,	88.8)
              (-0.000822828,	90.4)
              (-0.00151334,	    91.2)
              (-0.001308051,	87.6)
            };
            \addplot[only marks, mark=o, color=yellow, style={ultra thick}]
            coordinates{ % plot 4 data set
                (-0.000384712,  	79.2)
                (-0.000295034,	    88.8)
                (-1.72E-04,	        88.8)
                (-0.000185817,	    88)
                (-0.00025106,	    89.2)
                (-0.001087233,	    88.8)
                (-0.000171832,	    86.8)
                (-0.000276831,  	90)
                (-0.000213064,	    87.6)
                (-0.000422359,	    83.6)
                (-0.000889854,	    88)
                (-0.000219131,	    87.6)
                (-0.000195933,  	83.6)
                (-0.000196351,	    88.8)
                (-0.000278191,	    89.6)
            };

          \end{axis}
    \end{tikzpicture}
    \vspace*{1.5cm}
    \caption{\XPASC{}-\PPMI{} and accuracy results of runs across 15 seeds for SPAM. Different colors indicate different $\lambda$ values.
    Numbers in brackets must be multiplied by the subscript value, $10^{-3}$.}
    \label{fig:spamscatterppmi}
\end{minipage}
\end{figure}

%Scatter spouse
\begin{figure}
\centering
\begin{minipage}{.5\textwidth}
  \centering
    \begin{tikzpicture}[scale=0.65]
          \pgfplotsset{scale only axis,}
          \begin{axis}[
            xticklabels={1-(1.),1-(1.),1-(0.8), 1-(0.6), 1-(0.4), 1-(0.2), 1},
            xlabel=\XPASC{},
            ylabel=F1,
            legend entries={{$\lambda=0.0$}, {$\lambda=0.5$}, {$\lambda=1.0$}, {$\lambda=2.0$}, {$\lambda=4.0$}},
            legend columns=5,
            legend style={at={(0.98, 1.15)},anchor=north east},
            legend style={/tikz/every even column/.append style={column sep=0.3cm}}],
            xmin = -0.0012, xmax = 0,
            ymin = 0, ymax = 40,
          ]
            \addplot[only marks, mark=asterisk, color=blue]
            coordinates{ % plot 0 data set
              (-0.00018272, 	21.58)
              (-0.000190007,    19.23)
              (-0.000188094,	21.19)
              (-0.00021943,	    19.62)
              (-0.000237694,	18.83)
              (-0.000545937,	20.12)
              (-0.000269506,	20.04)
              (-0.000186887,	22.3)
              (-0.000235599,	21.77)
              (-0.000237325,	21.41)
              (-0.000213516,	19)
              (-0.000295914,	20.34)
              (-0.000563602,	19.34)
              (-0.00027986, 	22.29)
              (-0.000170526,	21.18)
              
            }; \label{plot_one}
            \addplot[only marks, mark=triangle, color=orange]
            coordinates{ % plot 05 data set
              (-0.000321763, 	26.64)
              (-0.000141867,	21.7)
              (-0.00019775, 	22.49)
              (-0.000149064,	20.02)
              (-0.00015572, 	20.93)
              (-0.000784988,	18.69)
              (-0.000119433,	31.84)
              (-0.000130067,	23.12)
              (-0.00012674,	    21.33)
              (-0.000173875,	21.41)
              (-0.000141558,	19.09)
              (-0.000142073,	19.48)
              (-0.000405044,	21.16)
              (-0.000324778,	22.73)
              (-0.000774708,	23.54)

            };
            \addplot[only marks, mark=diamond, color=green]
            coordinates{ % plot 1 data set
              (-0.000435078,	23.91)
              (-0.001071342,	23.12)
              (-7.97083E-05,	21.6)
              (-0.000170457,	20.04)
              (-0.000615902,	23.57)
              (-4.49521E-05,	20.85)
              (-9.82966E-05,	19.33)
              (-0.00072484,	    23.78)
              (-0.000158784,	25.9)
              (-0.000322242,	31.19)
              (-0.000385577,	18.27)
              (-0.000235039,	18.63)
              (-0.000205403,	22.22)
  
            };
             \addplot[only marks, mark=|, color=cyan, style={ultra thick}]
            coordinates{ % plot 2 data set
              (-4.90711E-05,	30.24)
              (-0.00077455,	    38.59)
              (-2.07641E-05,	15.27)
              (-1.03953E-06,	16.04)
              (-4.34771E-06,	15.43)
              (-4.83335E-06,	15.2)
              (-0.00018787,	    16.92)
              (-3.98066E-06,	15.61)
              (-4.43E-04,   	15.82)
              (-6.58064E-06,	16.94)
              (-7.14551E-06,	17.73)
              (-1.01855E-05,	39.45)
              (-9.22839E-06,	16.23)
              (-3.17916E-05,	14.52)
              (-1.88177E-06,	14.83)
     
            };
            \addplot[only marks, mark=o, color=yellow, style={ultra thick}]
            coordinates{ % plot 4 data set
              (-0.000287376,	14.84)
              (-7.92E-07,	    14.98)
              (-5.77581E-07,    15.05)
              (-7.11771E-07,	15.35)
              (-1.16049E-06,	13.8)
              (-0.000362566,	14.21)
              (-1.75685E-06,	15.09)
              (-5.59604E-07,	15.28)
              (-1.06633E-06,	15.85)
              (-8.32937E-07,	14.5)
              (-7.93322E-07,	15.88)
              (-4.49698E-07,	14.52)
              (-2.712E-06,  	15.07)
              (-5.9671E-07, 	15)
              (-8.89047E-07,	15.23)

            };

          \end{axis}
    \end{tikzpicture}
    \vspace*{1.5cm}
    \caption{\XPASC{}-\chis{} and accuracy results of runs across 15 seeds for SPOUSE. Different colors indicate different $\lambda$ values.
    Numbers in brackets must be multiplied by the subscript value, $10^{-3}$.}
    \label{fig:spousescatterchi2}
\end{minipage}%
\begin{minipage}{.5\textwidth}
      \centering
    \begin{tikzpicture}[scale=0.65]
          \pgfplotsset{scale only axis,}
          \begin{axis}[
            xticklabels={1-(8),1-(8), 1-(6), 1-(4), 1-(2), 1},
            xlabel=\XPASC{},
            ylabel=F1,
            legend entries={{$\lambda=0.0$}, {$\lambda=0.5$}, {$\lambda=1.0$}, {$\lambda=2.0$}, {$\lambda=4.0$}},
            legend columns=5,
            legend style={at={(0.98, 1.15)},anchor=north east},
            legend style={/tikz/every even column/.append style={column sep=0.3cm}}],
            xmin = -0.0009, xmax = 0,
            ymin = 0, ymax = 40,
          ]
            \addplot[only marks, mark=asterisk, color=blue]
            coordinates{ % plot 0 data set
              (-0.000278469,	21.58)
              (-0.000267144,    19.23)
              (-0.000271695,    21.19)
              (-0.000249715,    19.62)
              (-0.000269176,    18.83)
              (-0.000420924,    20.12)
              (-0.000275655,    20.04)
              (-0.0003186,      22.3)
              (-0.000265847,    21.77)
              (-0.000389098,    21.41)
              (-0.000265176,    19)
              (-0.000395335,    20.34)
              (-0.000474628,    19.34)
              (-0.000305385,    22.29)
              (-0.000296838,    21.18)

            }; \label{plot_one}
            \addplot[only marks, mark=triangle, color=orange]
            coordinates{ % plot 05 data set
              (-0.000560294,    26.64)
              (-0.00027089,     21.7)
              (-0.000276011,    22.49)
              (-0.00018569,     20.02)
              (-0.000301744,    20.93)
              (-0.000830238,    18.69)
              (-0.000162187,    31.84)
              (-0.000236762,    23.12)
              (-0.000188419,    21.33)
              (-0.000266444,    21.41)
              (-0.00021232,     19.09)
              (-0.000171918,    19.48)
              (-0.000481472,    21.16)
              (-0.000280412,    22.73)
              (-0.000634553,    23.54)

            };
            \addplot[only marks, mark=diamond, color=green]
            coordinates{ % plot 1 data set
              (-0.000128112,    23.91)
              (-6.6351E-07,     23.12)
              (-0.000132487,    21.6)
              (-0.000210214,    20.04)
              (-0.000593731,    23.57)
              (-5.8574E-05,     20.85)
              (-9.54979E-05,    19.33)
              (-0.000508144,    23.78)
              (-0.00016953,     25.9)
              (-0.000303503,    31.19)
              (-0.000378303,    18.27)
              (-0.000229678,    18.63)
              (-0.000287996,    22.22)

            };
             \addplot[only marks, mark=|, color=cyan, style={ultra thick}]
            coordinates{ % plot 2 data set
              (-6.37888E-05,    30.24)
              (-0.00067826,     38.59)
              (-1.04973E-05,    15.27)
              (-7.02572E-07,    16.04)
              (-3.26176E-06,    15.43)
              (-4.01214E-06,    15.2)
              (-7.98622E-05,    16.92)
              (-2.75332E-06,    15.61)
              (-0.000344537,    15.82)
              (-5.01955E-06,	16.94)
              (-5.07345E-06,    17.73)
              (-7.06942E-06,    39.45)
              (-6.53848E-06,    16.23)
              (-3.80297E-05,    14.52)
              (-1.28051E-06,    14.83)
 
            };
            \addplot[only marks, mark=o, color=yellow, style={ultra thick}]
            coordinates{ % plot 4 data set
              (-0.000128112,    14.84)
              (-6.6351E-07,     14.98)
              (-5.77581E-07,    15.05)
              (-4.69257E-07,    15.35)
              (-1.16049E-06,	13.8)
              (-0.000213518,    14.21)
              (-1.75685E-06,	15.09)
              (-6.96161E-07,    15.28)
              (-7.75485E-07,    15.85)
              (-6.3577E-07,     14.5)
              (-5.57217E-07,    15.88)
              (-4.98009E-07,    14.52)
              (-1.74E-06,       15.07)
              (-5.06739E-07,    15)
              (-6.14021E-07,    15.23)
            
            };

          \end{axis}
    \end{tikzpicture}
    \vspace*{1.5cm}
    \caption{\XPASC{}-\PPMI{} and accuracy results of runs across 15 seeds for SPOUSE. Different colors indicate different $\lambda$ values.
    Numbers in brackets must be multiplied by the subscript value, $10^{-4}$.}
    \label{fig:spousescatterppmi}
\end{minipage}
\end{figure}

\newpage 
\subsubsection{Magnitude of \XPASC{}}
The results show that the values for \XPASC{} are very small. 
This is a consequence of composition of the formula: 
On the one hand, the \chis{} and \PPMI{} values are low (often zero or close to zero) already and the final association value (Formula \ref{formula:asc}) becomes even smaller due to the subtraction.
On the other hand, the explainability values come from a probability distribution and therefore range between zero and one.
By multiplying these small values, the results become even smaller. 
To bring the \XPASC{} to a range with higher magnitude values, we experimented with the following normalization steps to calculate a scaled version of \XPASC{}:

\begin{itemize}
    \item scaling the range of the PPMI values between zero and one by using the normalized pointwise mutual information: \\
    \begin{equation}
        NPMI(f,z) = \frac{PMI(f,z)}{h(f,z)}
    \end{equation}
    where the pointwise mutual information in Equation \ref{PMI} is normalized by $h(f,z)$, with $h(f,z)$ being the joint self-information $ -log (P(f,z))$.
    
    \item scaling the explainability score to range between zero and one by normalizing the explainability of feature $f$ given instance $i$ by the maximum explainability value per instance: \\
    \begin{equation}
        S_{xp}(i,f) = \frac{D_{KL} \left ( P(i)||Q(i \setminus f) \right )}
    {max\left ( \left \{ \forall f'\in i \ | \ D_{KL} \left ( P(i)||Q(i \setminus f') \right ) \right\} \right )}
    \end{equation}
    where $P(i)$ is the model prediction for the entire instance and $Q(i \setminus f)$ or $Q(i \setminus f')$ is the model prediction for the instance with feature $f$ or $f'$ omitted. 
    
    \item scaling the distribution of both, explainability and association, to range between zero and one each by using MinMax scaling:\\
    \begin{equation}
        X_{std} = \frac{\left ( X  - X.min \right )}{\left ( X.max -  X.min \right )} \\
    \end{equation}
    \begin{equation}
        X_{scaled} =  X_{std} \times \left ( max  -  min)+ min \right )
    \end{equation}
    where $X$ is the array of all values (explainability or association) and $min/max$ indicate the minimum/maximum value of the array. 

\end{itemize}

\begin{figure}
\centering
\begin{minipage}{.5\textwidth}
  \centering
    \begin{tikzpicture}[scale=0.7]
          \pgfplotsset{scale only axis,}
          \begin{axis}[
            xticklabels={1-(8),1-(8), 1-(6), 1-(4), 1-(2), 1},
            xlabel=\XPASC{},
            ylabel=accuracy,
            legend entries={{$\lambda=0.0$}, {$\lambda=0.5$}, {$\lambda=1.0$}, {$\lambda=2.0$}, {$\lambda=4.0$}},
            legend columns=5,
            legend style={at={(0.98, 1.15)},anchor=north east},
            legend style={/tikz/every even column/.append style={column sep=0.3cm}}],
           xmin = -0.0009, xmax = 0,
            ymin = 78, ymax = 94,
          ]
            \addplot[only marks, mark=asterisk, color=blue]
            coordinates{ % plot 0 data set
              (-0.00077643,	    86)
              (-0.000591465,    87.6)
              (-0.000549923,    86)
              (-0.000635931,    85.2)
              (-0.000657075,    89.2)
              (-0.000432483,    91.2)
              (-0.000813277,    87.6)
              (-0.000601968,    86)
              (-0.000614098,    90)
              (-0.000629719,    90.4)
              (-0.00066138,     87.2)
              (-0.000625843,    88)
              (-0.000612872,    84.4)
              (-0.000492328,    88)
              (-0.000840937,    86.8)
            }; \label{plot_one}
            \addplot[only marks, mark=triangle, color=orange]
            coordinates{ % plot 05 data set
              (-0.000698687,    86.4)
              (-0.000500917,    87.2)
              (-0.00047420,     84.8)
              (-0.000498023,    87.2)
              (-0.000560134,    90)
              (-0.000362416,    91.6)
              (-0.000505191,    90)
              (-0.000509086,    89.2)
              (-0.000499741,    89.2)
              (-0.000560765,    88.8)
              (-0.000430352,    90.4)
              (-0.000462835,    85.6)
              (-0.000500781,    85.2)
              (-0.000567296,    88)
              (-0.000578167,    84)

            };
            \addplot[only marks, mark=diamond, color=green]
            coordinates{ % plot 1 data set
              (-0.000445865,    85.6)
              (-0.000377324,    86)
              (-0.000309691,    86.8)
              (-0.000290858,    90)
              (-0.000355848,    89.6)
              (-0.000264471,    88.8)
              (-0.000318033,    90)
              (-0.0003661,      90)
              (-0.000379205,    92.4)
              (-0.000396474,    89.6)
              (-0.000308644,    89.2)
              (-0.000309563,    86.9)
              (-0.000334466,    88.4)
              (-0.000215895,    88.4)
              (-0.000418465,    85.2)

            };
             \addplot[only marks, mark=|, color=cyan, style={ultra thick}]
            coordinates{ % plot 2 data set
              (-0.000209967,    89.2)
              (-0.000188653,    91.2)
              (-8.1231E-05,      90.8)
              (-9.15415E-05,    90)
              (-0.000151676,    90.4)
              (-8.2314E-05,     92)
              (-0.000135901,    91.6)
              (-0.000115564,    90)
              (-0.000152778,    90.4)
              (-8.93287E-05,    86.8)
              (-6.8579E-05,     90.4)
              (-0.000124832,    88.8)
              (-0.000120345,    90.4)
              (-0.000104936,    91.2)
              (-0.000166704,    87.6)
            };
            \addplot[only marks, mark=o, color=yellow, style={ultra thick}]
            coordinates{ % plot 4 data set
              (-4.29527E-05,    79.2)
              (-3.6593E-05,     88.8)
              (-2.02901E-05,    88.8)
              (-2.33453E-05,    88)
              (-3.61952E-05,    89.2)
              (-4.49353E-05,    88.8)
              (-2.57412E-05,    86.8)
              (-3.25776E-05,    90)
              (-3.67325E-05,    87.6)
              (-3.71847E-05,    83.6)
              (-7.3909E-05,     88)
              (-2.73598E-05,    87.6)
              (-2.65324E-05,    83.6)
              (-2.04182E-05,    88.8)
              (-3.99051E-05,    89.6)
            };
          \end{axis}
    \end{tikzpicture}
    \vspace*{1.2cm}
    %\caption{\XPASC{}-\chis{} and accuracy results of runs across 15 seeds for SPAM. Different colors indicate different $\lambda$ values.}
    \label{fig:spamscatterchi2old}
\end{minipage}%
\begin{minipage}{.5\textwidth}
  \centering
    \begin{tikzpicture}[scale=0.7]
          \pgfplotsset{scale only axis,}
          \begin{axis}[
            xlabel=\XPASC{},
            ylabel=accuracy,
            legend entries={{$\lambda=0.0$}, {$\lambda=0.5$}, {$\lambda=1.0$}, {$\lambda=2.0$}, {$\lambda=4.0$}},
            legend columns=5,
            legend style={at={(0.98, 1.15)},anchor=north east},
            legend style={/tikz/every even column/.append style={column sep=0.3cm}}],
           xmin = -0.0009, xmax = 0,
            ymin = 78, ymax = 94,
          ]
            \addplot[only marks, mark=asterisk, color=blue]
            coordinates{ % plot 0 data set
(0.07288869, 86)
             (0.073310833, 87.6)
             (0.073674127, 86)
             (0.073384411, 85.2)
             (0.07287829, 89.2)
             (0.077876618, 91.2)
             (0.074059518, 87.6)
             (0.076273, 86)
             (0.075955305, 90)
             (0.075858642, 90.4)
             (0.074877916, 87.2)
             (0.076146386, 88)
             (0.074329349, 84.4)
             (0.072422861, 88)
             (0.072920243, 86.8)
           }; \label{plot_one}
           \addplot[only marks, mark=triangle, color=orange]
           coordinates{ % plot 05 data set
             (0.073582568,    86.4)
             (0.07400107,    87.2)
             (0.075068113,     84.8)
             (0.076398716,    87.2)
             (0.071563463,    90)
             (0.073261138,    91.6)
             (0.074729161,    90)
             (0.073925821,    89.2)
             (0.072162475,    89.2)
             (0.070945428,    88.8)
             (0.071489901,    90.4)
             (0.074461859,    85.6)
             (0.074327809,    85.2)
             (0.073782708,    88)
             (0.075180753,    84)

           };
           \addplot[only marks, mark=diamond, color=green]
           coordinates{ % plot 1 data set
             (0.07239887,    85.6)
             (0.072420102,    86)
             (0.075563602,    86.8)
             (0.073380943,    90)
             (0.074386107,    89.6)
             (0.074473095,    88.8)
             (0.075266533,    90)
             (0.073849545,      90)
             (0.074501668,    92.4)
             (0.074453031,    89.6)
             (0.07825229,    89.2)
             (0.072651307,    86.9)
             (0.074484639,    88.4)
             (0.077830095,    88.4)
             (0.077042281,    85.2)

           };
            \addplot[only marks, mark=|, color=cyan, style={ultra thick}]
           coordinates{ % plot 2 data set
             (0.073427024,    89.2)
             (0.073702199,    91.2)
             (0.074288692,      90.8)
             (0.071648874,    90)
             (0.07862129,    90.4)
             (0.072568073,     92)
             (0.073969915,    91.6)
             (0.077334255,    90)
             (0.077887222,    90.4)
             (0.074168742,    86.8)
             (0.074201442,     90.4)
             (0.07423619,    88.8)
             (0.075642946,    90.4)
             (0.078179345,    91.2)
             (0.075195143,    87.6)
           };
           \addplot[only marks, mark=o, color=yellow, style={ultra thick}]
           coordinates{ % plot 4 data set
             (0.075778532,    79.2)
             (0.075808246,     88.8)
             (0.074799782,    88.8)
             (0.076427146,    88)
             (0.075960129,    89.2)
             (0.074979845,    88.8)
             (0.078931766,    86.8)
             (0.075510411,    90)
             (0.07387422,    87.6)
             (0.076068078,    83.6)
             (0.077150978,     88)
             (0.073489315,    87.6)
             (0.078192119,    83.6)
             (0.07387294,    88.8)
             (0.077235603,    89.6)
            };
          \end{axis}
    \end{tikzpicture}
    \vspace*{1.2cm}
    %\caption{\textbf{Normalized} \XPASC{}-\chis{} and accuracy results of runs across 15 seeds for SPAM. Different colors indicate different $\lambda$ values.}
    \label{fig:spamscatterchi2new}
\end{minipage}
\caption{Original and \textbf{normalized} \XPASC{}-\chis{} and accuracy results of runs across 15 seeds for SPAM. Different colors indicate different $\lambda$ values.
Numbers in brackets must be multiplied by the subscript value, $10^{-4}$.} 
\label{fig:comparespam}
\end{figure}
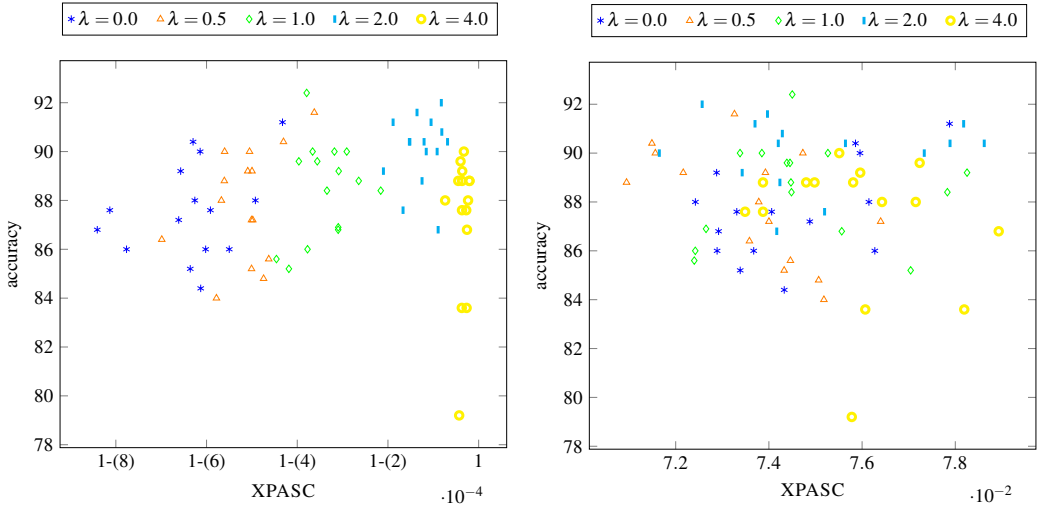

This results in larger \XPASC{} values, but normalizing and scaling the values can discard useful information. 
To investigate if there is an information loss due to the normalizing of the score, we compute another scatter plot that depicts the performance and generalization per run and seed for SPAM after applying the above mentioned steps to \XPASC{} (see Figure \ref{fig:comparespam}).
As one can clearly see, the individual characteristics are no longer as distinctly recognizable in the normalized version of \XPASC{} (right plot) as they had been before (left plot). 
While the values for the individual \KnowMAN{} models used to be clearly separated from each other, the normalization has mixed them up.
We conclude that important information is lost due to the normalization of the values.
For this reason we accept the small values of the original formula in order to be able to optimally represent all information.

%Scatter spam no normalization

\newpage

\subsection{Qualitative Feature Analysis}
To confirm the functionality of \XPASC{} qualitatively we took a close look at the data and their linguistic features to find instances that illustrate \XPASC{} and its components. 

First, we examined the association matrices to verify that they reflect the actual association of features, classes and labeling functions.
See Tables \ref{tab:lfs} and \ref{tab:lfsspouse} for the top five association values (for both \chis{} and \PPMI{}) of labeling functions and features for SPAM and SPOUSE. Because of the larger number of labeling functions we did not conduct this analysis for TREC and IMDb. 

With regard to the \chis{}-based association, the association between the labeling functions and the features can be understood easily.
Many of the features most associated with their labeling function would also be considered very relevant by a human. In some cases, the features are even part of the pattern of the labeling function. 
For the \PPMI{}-based association the top features contain parts of the patterns only rarely.
Note, that \PPMI{} is very sensitive to features that occur only once.
These features obtain very high \PPMI{}-based association scores because they are observed exclusively with a single class or labeling function.
However, since these features all have a low frequency (occur only once), this is not a problem for the calculation of \XPASC{}.
Still, one can find words that are part of common expressions together with the keyword of the labeling function, e.g.  ``{}ALBUM''{} is associated with  ``{}my''{}. 
For SPOUSE one can find a lot of names among the \PPMI{}-based association. 
To figure out if persons are married it is plausible that persons names are considered a lot.  

\begin{table}
    \centering
    \small
    \caption{Top 5 labeling function related features SPAM.}
    \vspace*{0.2cm}
    \begin{tabularx}{\textwidth}{X|c|X|X}
    \hline
       \textbf{labeling function}  & \textbf{class} & \textbf{\chis{}} &  \textbf{\PPMI{}} \\ \hline

         keyword ``{}my''{} & SPAM  & my, channel, Hey, MY, probably & Cyphers, ALBUM, TOMORROW,  READING, tvcmcadavid.weebly\\ \hline
         
         keyword ``{}subscribe''{} & SPAM &  me, subscribe, to, subscribers, subscribe & Del, Rey, Drake, Macklemore, Pink \\ \hline
         
         link & SPAM &  V, \textbackslash, STYLE, fight, 'http://youtu.be/9bZkp7q19f0' & scrubs, ./r, MontageParodies, AND, OTHER\\ \hline
         
         keyword ``{}Please''{} & SPAM & plz, please, Please, PLEASE, school & \textbf{\textendash}, \textbf{\dag},\textbf{\textbar} , PLZZ, supporters\\\hline
         
         keyword ``{}song''{} & HAM &  song, This, songs, fun, Best & spare, upcoming, uk, rapper.please, worries  \\ \hline
         
         regex ``{}Check out''{} & SPAM &  out, this, on, Check, video &  act, retain, delightful, system, rhythm \\ \hline
        
         short comment & HAM & out, this, on, \_, and & BR, sparkling-heart emoji, wonderful, LOST?, heart emoji\\ \hline
         
         has person & HAM & Katy, Perry, Official, Charlie, Eminem & belle, chanson, lost?, clean, Eminem \\ \hline 
         
         polarity $>$ 0.9 & HAM &  best, photo, Oppa, Yeah, Best &  MOVES, MAKES, MEH, SMILE, EVER\\ \hline
         
         subjectivity $>=$ 0.5 & HAM & only, views, YouTube:, love, out  & Driveshaft, YEAH, Crazy, Flow, Ill\\ \hline
    \end{tabularx}
    \label{tab:lfs}
    \vspace*{0.4cm}
\end{table}
\begin{table}[hbt!]
    \centering
    \small
    \caption{Top 5 labeling function related features for SPOUSE.}
    \vspace*{0.2cm}
    \begin{tabularx}{\textwidth}{X|c|X|X}
    \hline
       \textbf{labeling function}  & \textbf{class} & \textbf{\chis{}} &  \textbf{\PPMI{}} \\ \hline

         keyword ``{}husband/wife''{} & SPOUSE  & married, son, wife, husband, boyfriend & Peck, Veronique, glitters, Sweeting's, Body\\ \hline
         
         keyword ``{}husband/wife''{} left window & SPOUSE &  written, wife, husband, conspiring, tunnel & Guys, Running, Role, wrestle, Off \\ \hline
         
         same last name & SPOUSE & son, wife, husband, afternoon, daughter & Dudley, Hales, facilitating, Thomson, M\&F \\ \hline
         
         keyword ``{}married''{} & SPOUSE &  married, relationship, 2007., who, trainer & wakes, Gordon-Levitt, rowing, Regardless, minorities  \\ \hline
         
         familiar relationship & NO SPOUSE &  son, wife, husband, sister, daughter &  Fire, window, forcing, ribs, Claims \\ \hline
        
         familiar relationship left window & NO SPOUSE & on, husband, half-brother, daughter, mother & hopelessness, hairdresser, sponsoring, Tailyour, Commandant \\ \hline
         
         regex other relationship & NO SPOUSE & husband, boyfriend, planted, Gamble, David & Rita, Ora, Grimshaw, Zeinat, ordinary \\ \hline 
         
         known spouses from database & SPOUSE & Martin, denim, afternoon, Gwyneth, Paltrow & lacing, Amicable, exes, shifts, Basinger \\ \hline
         
         last name known spouses & SPOUSE & Mara, mirror, tank, Paltrow, beauties  & reported, Radar, spousal, grocery, head-on \\ \hline
    \end{tabularx}
    \label{tab:lfsspouse}
    \vspace*{0.5cm}
\end{table}
\begin{table}
    \small   
    \centering
    \caption{Top 10 class related features.}
    \vspace*{0.2cm}
    \begin{tabularx}{\textwidth}{X|c|X|X}
    \hline
        data set & class & \chis{} & \PPMI{} \\ \hline \hline
        \multirow{5}{*}{\textbf{SPAM}} & SPAM & subscribe, please, check, out, my, channel, song, Please, on, Check & beat, losing, ideas, lies, history, Driveshaft, YEAH, hot, Beats!!, STTUUPID \\ \cline{2-4}
        & HAM & subscribe, please, check, out, my,  will, channel, song, Please, on & Drake, Macklemore, Pink, countless, inspire, FYI, freedom, speech, Lil, uploaded \\ \hline \hline
        \multirow{17}{*}{\textbf{TREC}} & description & cleaveland, cavaliers, monarchy, added, quisling, repossesion, butcher, handful, spine, currency & per, chicken, dog, capital, university, black, island, san, south, west \\ \cline{2-4}
        & entity & sailed, talk-show, lends, surroundings, thalia, shakespearean, shylock, airforce, compiled, won-lost & leader, animals, words, father, christmas, held, nicknae, law, only, john\\ \cline{2-4}
        & human being & builds, resistance, odor, auh2o, mccain, rifleman, lai, malawi, zebulon, pike & god, square, mile, gas, strip, court, basketball, nationality, rock,month \\ \cline{2-4}
        & abbreviation & olympic, original, committee, aids, manufacturer, cpr, abbreviation, p.m., trinitrotoluene, equipment & monarchy, added, quisling, puerto, rico, repossession, butcher, handful, spine, 's\\ \cline{2-4}
        & location & aborigines, adventours, tours, photosynthesis, makeup, erykah, badu, m, ayer, bend & said, so, letter, kennedy, bridge, human, nixon, no, river, his \\ \cline{2-4}
        & numeric value & 56-game, streak, graffiti, quilting, iran-contra, deere, tractors, cherubs, puerto, rico & square, strip, court, jackson, basketball, numbers, university, john, show \\ \hline \hline
        \multirow{5}{*}{\textbf{SPOUSE}} & SPOUSE & married, son, wife, husband, boyfriend, …, young, family, sister, daughter & ringing, Sweeting's, Body, Cutting, Crap, Australians, marches, splashes, Kingi \\ \cline{2-4}
        & NO SPOUSE & married, son, wife, husband, boyfriend, …, young, family, younger, sister & ordinary, rank-and-file, handpicked, Abu, Bakr, al-Baghdadi, L.L., J.'s, trendy \\ \hline \hline
        \multirow{7}{*}{\textbf{IMDB}} & negative & Instead, back, character, too, much, does, entire, cast, So, bizarre & Zwarts, Fredrikstad, Hilarios!10, wawa, CONSIDERING, Hobb's, Smooch, Investigative, belly-dancers, retirony \\ \cline{2-4}
        & positive & writing, It's, most, drag, you, us, won, Oscar, those, endless & Culpability, Package, slip-ups, AARP, Symona, Boniface, Lorch, Lynton, Tyrrell, Heinie \\  \hline
    \end{tabularx}
    \vspace*{2cm}
    \label{tab:classassoc}
\end{table}

The association of features with the classes is not as intuitive as for the labeling functions. 
For \chis{}-based association the top ten features are identical for the classes for SPAM and almost identical for SPOUSE.
Note that the \chis{}-association only expresses which features were particularly relevant for determining the class. However, the correlation of these features with the class can be both positive (the feature is strongly associated because it gives a clue to the correct class) and negative (the feature is strongly associated because it gives a clue to distinguish it from other classes). 
The association with each class for the TREC data set is more intuitive in some cases. 
Words like ``{}odor''{} or ``{}malawi''{} are associated with the class ``{}human being''{}, ``{}cpr''{} or ``{}p.m.''{} with ``{}abbreviation''{}, what is plausible.
On the other hand words like ``{}make up''{} are associated with ``{}location''{}. 
Since only one gold class could be chosen by the annotators during the labeling of the data set, it is likely that some words would also be suitable for another class that had been in the set of suitable classes associated with the instance.
For IMDb the features are different for both classes and very intuitive for a human. Features like ``{}bizarre''{} or ``{}Oscar''{} clearly point to a certain sentiment.

For \PPMI{}-based association the top ten features are different for all data sets. 
Again this ranking is not easily interpretable for a human, but reflects the association and co-occurrence in the weakly supervised data sets. 
Because of the sensitivity to rare words, we found many features with the same association score, and the top n features in Table \ref{tab:classassoc} are therefore only an excerpt.

\begin{figure}
    \centering
    \includegraphics[scale=0.4]{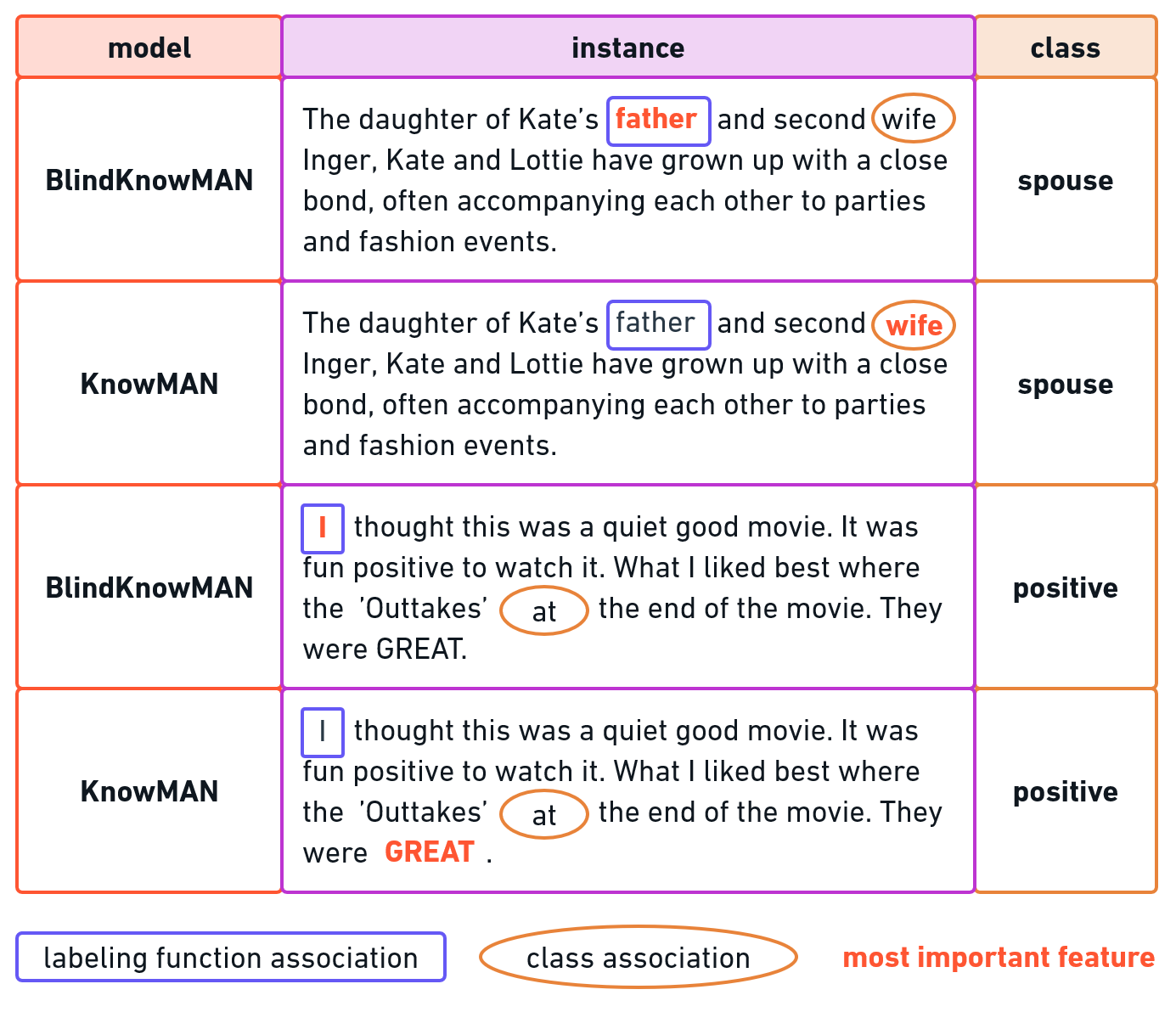}
    \vspace*{0.8cm}
    \caption{Examples from SPOUSE and IMDb where the feature with the highest explainability is shifted from a misleading labeling function towards the correct class. Association is based on \chis{} and \KnowMAN{} uses $\lambda=4$.}
    \label{fig:ex1}
\end{figure}

Next, we looked for instances that confirm the functionality of \XPASC{} and \KnowMAN{}. Therefore, we compared \bKM{} and \KnowMAN{} with $\lambda=4$.
The ultimate and most challenging requirement for the models would be the following:
Shift the focus from features that are associated with a deceptive labeling function towards features that are associated with the correct class. 
Two kinds of information need to be found for that goal: 
i) a feature that is very important for the classification (has a very high explainability score) and is associated most with a misleading labeling function, pointing to the wrong class, in \bKM{} and 
ii) the \KnowMAN{} model is able to shift the highest explainability to a feature that is not associated with the misleading labeling function anymore, but with the correct class. 
Thus in the \KnowMAN{} model the most important feature should be associated with the correct class most.
For example in Figure \ref{fig:ex1} the first instance should be classified as holding a \textit{spouse} relation. 
The most important feature for \bKM{} is \textbf{father}, what actually would lead to the classification of \textit{no spouse}. 
The \KnowMAN{} model achieves the shift to the feature \textbf{wife}, that is a better indicator for the spouse relation.
The second example in Figure \ref{fig:ex1} is drawn from IMDb. 
The review is \textit{positive} and the feature '{}I'{} is associated with the negative class. 
\KnowMAN{} manages to shift the focus to the feature '{}GREAT'{}, what is a better indicator for a positive sentiment.

We also measured this shift of the most important feature quantitatively.
For SPAM, we found a shift is achieved for 12 instances and for SPOUSE 267 instances (both on average across seeds). 
For IMDb the model manages to shift the misleading feature to the correct one for 7 instances. This comparison always refers to \bKM{} and \KnowMAN{} with $\lambda=4$.

\begin{figure}
    \centering
    \includegraphics[scale=0.4]{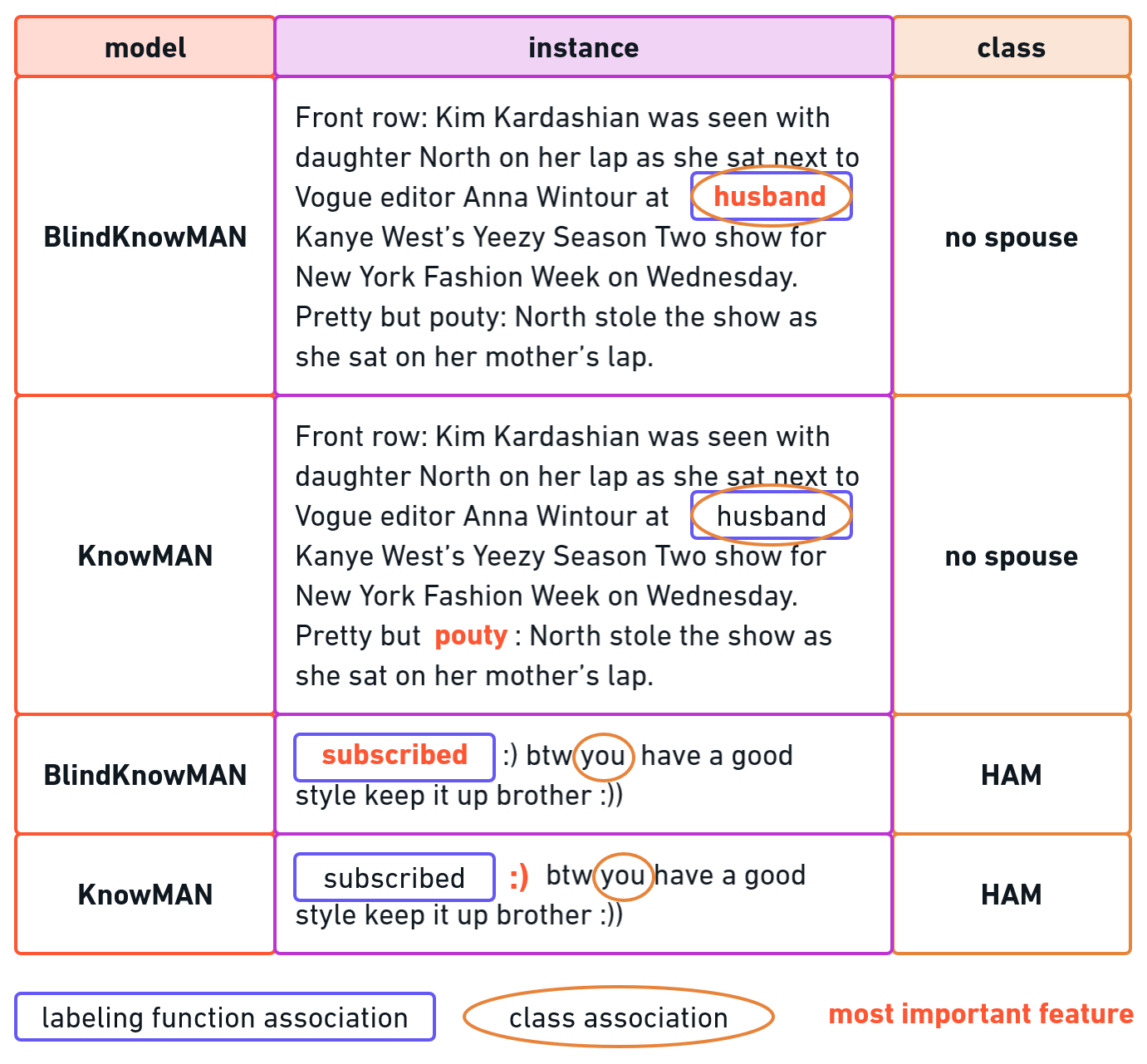}
    \vspace*{0.8cm}
    \caption{Examples from SPAM and SPOUSE. Highest explainability is shifted from features that are associated with the labeling function to other features. Association is based on \chis{} and \KnowMAN{} uses $\lambda=4$.}
    \label{fig:ex2shift}
\end{figure}

A better generalization can also be achieved if the focus is shifted from the misleading feature to another feature that is not associated with the correct class, but at least is no longer associated with the labeling function. 
See Figure \ref{fig:ex2shift} for examples that illustrate this. 
The first example, again from SPOUSE, expresses a \textit{no spouse} relation, but the most important feature for \bKM{} is \textbf{husband}. Shifting the focus to another word - \textbf{pouty} - \KnowMAN{} is able to assign the correct label.
The second example is comes from SPAM, where \bKM{} considers the most important feature as \textbf{subscribed} for an instance that actual belongs to the \textit{HAM} class. 
Since this is misleading, \KnowMAN{} focuses on the emoticon in the instance and assigns the correct label.

In addition, we noticed in the linguistic feature-based analysis that the weak labels for SPOUSE are very noisy and imprecise. 
We found many instances where a human annotator would have assigned another class than the labeling functions assigned. 

Overall, the quantitative results can confirm our findings of the qualitative analysis. 
The \KnowMAN{} architecture is able to increase generalization and \XPASC{} is a good indicator for the generalization ability of models.
\section{Conclusion}
We presented \XPASC{}, a novel score to measure generalization for weakly supervised models. 
Our extensive analysis shows that \XPASC{} is able to reflect the generalization of models given a dataset and the labeling functions used to perform weak supervision. 
In addition, we studied the adversarial approach \KnowMAN{}, designed to enable the control of generalization in weakly supervised models. 
We confirmed the hypothesis that the architecture is able to control the shift from labeling functions to other signals by a hyperparameter.
We also showed that performance and generalization do not relate one-to-one and it has to be decided based on the task, dataset and model, which degree of generalization is desired.
\XPASC{} can be used with any pre-trained weakly supervised model, a dataset and its set of applied labeling functions. 
Assuming that many neural models, designed to work with noisy weakly supervised data, are complex and thus suffer from overfitting, \XPASC{} can serve as an indicator for their ability to fit unseen data.
In general the core components of \XPASC{}, explainability and association, are interchangeable, what makes the score flexible in practice.

\section*{Competing interests declaration}
Competing interests: The authors declare none.

\section*{Acknowledgements}
This research was funded by the WWTF through the project "Knowledge-infused Deep Learning for Natural Language Processing" (WWTF Vienna Research Group VRG19-008).

\bibliographystyle{nlelike}
\bibliography{new_refs}

\label{lastpage}

\end{document}